\documentclass[11pt]{article}
\usepackage[preprint]{acl}
\usepackage{times}
\usepackage{latexsym}
\usepackage[T1]{fontenc}
\usepackage[utf8]{inputenc}
\usepackage{microtype}
\usepackage{inconsolata}
\usepackage{graphicx}
\usepackage{makecell}
\usepackage{url}
\usepackage{paracol} 
\usepackage{amsmath}
\usepackage{algorithm, algpseudocode}

\usepackage{graphicx}
\usepackage{subcaption}	
\usepackage{multirow}
\definecolor{bg}{RGB}{176,226,255}
\definecolor{bonus_green}{RGB}{0,100,0}

\usepackage{tabularx}
\usepackage[table,xcdraw]{xcolor}
\usepackage{booktabs}
\usepackage{amssymb}
\usepackage{xcolor}
\usepackage[most]{tcolorbox}
\usepackage{listings}
\tcbuselibrary{listings}
\usepackage[utf8]{inputenc}
\usepackage{microtype}
\usepackage{inconsolata}
\usepackage{graphicx}
\usepackage{natbib}
\usepackage{amsmath}
\usepackage{relsize}
\usepackage{hyperref}
\usepackage{titling}

\title{ADORA: Training Reasoning Models with Dynamic Advantage Estimation on Reinforcement Learning}
\author{Qingnan Ren$^{1}$ \quad Shiting Huang$^{1}$ \quad Zhen Fang$^{1}\thanks{\quad Equal Contribution}$\\
\textbf{Zehui Chen}$^{1}$ \quad \textbf{Lin Chen$^{1}$} \quad \textbf{Lijun Li$^{2}$}  \quad \textbf{Feng Zhao$^{1}\thanks{\quad Corresponding author}$} \\
$^{1}$University of Science and Technology of China \quad $^{2}$Shanghai AI Laboratory
}

\begin{document}
\maketitle

\begin{abstract}
Reinforcement learning has become a cornerstone technique for developing reasoning models in complex tasks, ranging from mathematical problem-solving to imaginary reasoning.
The optimization of these models typically relies on policy gradient methods, whose efficacy hinges on the accurate estimation of an advantage function.
However, prevailing methods typically employ static advantage estimation, a practice that leads to inefficient credit assignment by neglecting the dynamic utility of training samples over time.
This limitation results in suboptimal policy updates, which in turn manifest as slower convergence rates and increased learning instability, as models fail to adapt to evolving sample utilities effectively.
To address this problem, we introduce \textbf{ADORA} (\textbf{A}dvantage \textbf{D}ynamics via \textbf{O}nline \textbf{R}ollout \textbf{A}daptation), a novel framework for policy optimization.
ADORA dynamically adjusts the advantage function's weighting by adaptively categorizing training data into temporarily advantageous and disadvantageous samples, based on their evolving utility during online model rollouts.
This tailored data differentiation strategy allows ADORA to be seamlessly integrated into existing policy optimization algorithms without significant architectural modifications, enabling the policy to prioritize learning from more informative experiences and thereby achieve more efficient policy updates.
Extensive evaluations across diverse model families and varying data scales demonstrate that ADORA is a robust and efficient framework. It significantly enhances long reasoning in both geometric and mathematical tasks, consistently achieving notable performance gains without requiring sensitive hyperparameter tuning.
\end{abstract}
\section{Introduction}
\label{eq:Introduction}
Recent developments of reasoning models, exemplified by R1 \cite{guo2025deepseek}, have expanded the scope of large language models (LLMs) into a reinforcement learning (RL) based paradigm.
By introducing long chain-of-thought (CoT) reasoning, these models can achieve effective test-time scaling and generate more sophisticated reasoning patterns, including verification, reflection, and backtracking \cite{guo2025deepseek,xie2025logic}. 
This capability is further internalized within the model through RL, which enhances generalization and enables it to address complex real-world problems, such as math \cite{liu2025understanding}, agent \cite{feng2025group}, and visual reasoning \cite{wang2025vl,huang2026vision,lab2025safework}.
Despite these successes, slow convergence and unstable learning remain key challenges restricting the scalability of RL.

To enable scalable RL, it is crucial to efficiently utilize samples to achieve both fast convergence and stable learning.
However, existing methods such as Proximal Policy Optimization (PPO) \cite{schulman2017proximal} and Group Relative Policy Optimization (GRPO) \cite{zhang2025grpo} assume that the informativeness of each training example remains constant throughout policy optimization, ignoring the dynamic nature of learning. This results in diminished learning gains from individual samples, slower convergence, and a greater demand for training iterations and data to achieve an acceptable performance level, thereby significantly limiting both training efficiency and the ultimate performance potential of reinforcement learning. To address this issue, our key insight is that \textbf{a sample's advantage should evolve alongside the policy}. Specifically, as the model is trained and the policy improves, the learning signal provided by the same example changes over different training iterations. Some samples may provide significant learning opportunities at certain stages, while others may involve concepts that are either already mastered or beyond the model’s current capacity to learn effectively. Treating all samples with uniform importance or with pre-defined static weights fails to leverage this dynamic utility, potentially leading to suboptimal learning trajectories and inefficient data use, as also noted by observations that current methods lack robust mechanisms for handling samples of varying utility during training \cite{ye2025limo}.
Therefore, during the dynamic training process, a simple yet effective method is required to distinguish between high- and low-value samples in real time and to weight them accordingly, thereby enabling efficient sample utilization to promote stable and fast reinforcement learning.
Motivated by these patterns and our key insight, we propose \textbf{ADORA} (\textbf{A}dvantage \textbf{D}ynamics via \textbf{O}nline \textbf{R}ollout \textbf{A}daptation), a novel and unified RL framework designed to dynamically calibrate advantage estimation for both LLMs and
VLMs. ADORA categorizes training data into Temporarily Advantageous Samples (TAS) and Temporarily Disadvantageous Samples (TDS) based on the model's rollout performance under a predefined data differentiation strategy. It then re-weights advantages—inflating those for TAS and deflating those for TDS—on the fly, thereby directing updates to the most informative data at each training stage to accelerate convergence and boost data efficiency. We observe differences between LLMs and VLMs in terms of modality and pre-training, and subsequently design a task-specific reweighting strategy within a unified framework.

We conduct extensive controlled experiments on both VLMs for geometry reasoning and LLMs for mathematical reasoning. 
Our experiments cover a wide range of architectures (Dense and MoE) and model families, including Llama-3, Mistral, DeepSeek, and InternVL. 
Empirically, ADORA significantly improves long chain-of-thought reasoning and task generalization. For instance, on the Qwen-7B-base model, ADORA achieves an average of 3.4 percentage points improvement over vanilla GRPO on math tasks. For VLMs, using fewer than 2,000 samples and no task-specific cold-start, the Qwen2.5-VL-7B model achieves 73.5\% accuracy on MathVista with ADORA.

Our key contributions and findings include:
\begin{itemize}
    \item \textbf{The ADORA framework}: We propose a simple, elegant, and efficient method for dynamically calibrating advantage estimation weights in RL based on live rollout statistics. 
    \item  \textbf{Task-specific differentiation strategies}: We design and validate distinct strategies for distinguishing TAS and TDS across different reasoning domains, consistently demonstrating improvements over vanilla-GRPO.
    \item \textbf{Comprehensive empirical analysis}: Extensive experiments are conducted to statistically evaluate ADORA across multiple dimensions, including training dynamics and thinking patterns, thereby offering insights into its underlying mechanisms. We further provide detailed ablation studies demonstrating that ADORA is robust to hyperparameter variations, effective under different advantage criteria, and applicable to diverse RL algorithms, establishing it as a stable and generalizable framework.
\end{itemize}
\section{Related Works}
\paragraph{Curriculum Learning.}
The core idea of Curriculum Learning (CL) \cite{bengio2009curriculum,elman1993learning} is to present training samples in a meaningful order, typically from easy to hard, to enhance learning efficiency and generalization.
Several variants have been proposed.  \cite{kumar2010self}dynamically selects easier samples based on the model's current prediction loss, thereby implementing an easy-to-hard training schedule.   \cite{matiisen2019teacher}introduces a teacher-student framework where the teacher selects sub-tasks demonstrating the fastest learning progress for the student, guided by the student's learning curve. More recently,  \cite{wang2025dump} dynamically adjusts sampling probabilities across different data distributions to achieve an adaptive training schedule.  \cite{deng2025boosting} proposed a three-stage reinforcement learning approach employing a progressive difficulty reward mechanism to optimize RL training.  \cite{wen2025sari} utilizes a two-stage curriculum-guided training.
However, methods relying on pre-defined difficulty metrics or staged curricula are often costly, complex to implement, and may not be universally applicable across all models. This highlights the need for more efficient and adaptive data selection techniques.

\paragraph{Reinforcement Learning for Reasoning in LLMs and VLMs.}
Leveraging GRPO, DeepSeek-R1 \cite{guo2025deepseek} demonstrated significant improvements in reasoning capabilities through rule-based reward reinforcement learning (RL), often accompanied by the emergence of reflection tokens and an increase in the length of Chain-of-Thought (CoT) \cite{wei2022chain} responses. Subsequent research has extensively applied R1-style rule-based RL to LLMs \cite{xie2025logic,zeng2025simplerl,yan2025learning} and VLMs \cite{shen2025vlm,li2025think,meng2025mm}.
On one hand, efforts have focused on optimizing GRPO. For instance,  \cite{yu2025dapo}introduced decoupled clipping and dynamic sampling strategies, among other techniques, to enhance RL training stability and efficiency for long-chain reasoning tasks.  \cite{zhang2025grpo}incorporated mechanisms such as length-aware accuracy rewards and error penalties.
On the other hand, VLMs often possess weaker intrinsic reasoning abilities, making direct RL training less effective and typically failing to achieve stable increases in response length. This has led to strategies such as cold-starting with large-scale data \cite{huang2025vision} or multi-stage training, sometimes beginning with text-only data to enhance model capabilities \cite{peng2025lmm}. 

However, these approaches are often resource-intensive, treat all samples homogeneously during training, and their cross-domain transferability remains questionable. In contrast, ADORA dynamically assesses whether samples are \emph{advantageous} or \emph{disadvantageous} to scale the advantage estimation signal in real-time, which allows the model to prioritize high-potential instances and accelerates the emergence of reasoning capabilities from scratch.

\section{Method}

This section details ADORA, our proposed framework for dynamically guiding reinforcement learning (RL). 
We begin with a brief review of prevailing RL algorithms in Section~\ref{sec:Preliminaries}, providing insights into the limitations of static advantage estimation.
Building on this analysis, we then present ADORA in Section~\ref{sec:ADORA}, which dynamically re-weights the contribution of training samples, and demonstrate its adaptability across both weaker and stronger reasoning models.

\subsection{Preliminaries}
\label{sec:Preliminaries}
The generation process of a language model can be modeled by a conditional policy $\pi_\theta$, which produces an output sequence $\mathbf{o}$ given an input $\mathbf{q}$.
At each step $t$, the model samples a token $o_t$ from the vocabulary according to the distribution $\pi_\theta(o_t\mid q, o_{<t})$.
The quality of a generated response $\mathbf{o}$ for a given input $\mathbf{q}$ can be evaluated by a reward function $R(\mathbf{q}, \mathbf{o})$.  
To align the model with desired behaviors, RL fine-tuning maximizes the expected reward while constraining the policy to remain close to a reference model $\pi_{ref}$.
The optimization objective is: 

\vspace{-0.5em}
{\small
\begin{equation}
\begin{aligned}
\mathcal{J}(\theta)
&=
\mathbb{E}_{\mathbf{q} \sim p_{\mathcal Q},\;\mathbf{o} \sim \pi_\theta(\cdot|\mathbf{q})}
\\
&\quad
\!\Big[R(\mathbf{q},\mathbf{o})
- \beta\,\mathbb{D}_{\mathrm{KL}}\!\big(\pi_\theta(\cdot|\mathbf{q})\|\pi_{\text{ref}}(\cdot|\mathbf{q})\big)\Big]
\label{eq:rl_objective}
\end{aligned}
\end{equation}
}

\noindent here, $p_{\mathcal{Q}}$ is the distribution of input queries, and $\beta$ controls the strength of KL regularization.

Prevailing RL approaches, such as PPO \cite{schulman2017proximal}, optimize the objective in Equation~\ref{eq:rl_objective} using policy gradient methods.
Unlike PPO, which typically relies on Generalized Advantage Estimator \cite{schulman2015high}, Group Relative Policy Optimization (GRPO) \cite{zhang2025grpo} avoids a separate value network by computing sample-wise advantages directly from normalized rewards across a group of rollouts.
Specifically, let $\mathcal{D}=\{(q,a)\}$ represent a dataset of question–answer pairs.
For each sample $q$, a group of $G$ individual responses $\{o_i\}_{i=1}^G$ is generated the old policy $\pi_{\theta_{\text{old}}}$ and assigned rule-based rewards $\{R_i\}_{i=1}^G$.
The estimated advantage $\hat{A}_{i,t}$ is identical across all tokens within a response, which is derived from the group rewards as:

\vspace{-0.5em}
{\small
\begin{equation}
\hat{A}_{i,t}=\hat{A}_{i}=\frac{r_i-\operatorname{mean}(\{R_i\}_{i=1}^G)}{\operatorname{std}(\{R_i\}_{i=1}^G)}
\label{eq:rl_adv}
\end{equation}
}

GRPO adapts PPO's clipped objective to optimize Equation~\ref{eq:rl_objective} using the group-level advantage estimate:

\vspace{-0.5em}
{\small
\begin{equation}
\begin{aligned}
\mathcal{J}_{\mathrm{GRPO}}(\theta) 
&= \mathbb{E}_{(q,a)\sim\mathcal{D},\{o_i\}_{i=1}^G\sim\pi_{\theta_{\mathrm{old}}}(\cdot|q)} \\
&\quad \Bigg[ \frac{1}{G}\sum_{i=1}^G \frac{1}{|o_i|} \sum_{t=1}^{|o_i|} \Big( \min ( \rho_{i,t}(\theta)\hat{A}_{i,t}, \\
&\qquad \mathrm{clip}\big(\rho_{i,t}(\theta), 1-\epsilon, 1+\epsilon\big)\hat{A}_{i,t} \Big) \\
&\qquad - \beta D_{\mathrm{KL}}(\pi_\theta \| \pi_{\mathrm{ref}}) \Big) \Bigg]
\end{aligned}
\end{equation}
}

\noindent where $\rho_{i,t}(\theta)=\frac{\pi_\theta(o_{i,t}\mid q,o_{i,<t})}{\pi_{\theta_{\mathrm{old}}}(o_{i,t}\mid q,o_{i,<t})}$ is the importance weight.

Crucially, the per-sample advantage is computed from rewards and remains static throughout an epoch or even the entire training process for that sample in those algorithms. Under optimization with static advantage estimates, all successful rollouts are \textbf{treated equally regardless of their informativeness}, which limits the adaptability of such methods to the model’s evolving capabilities as discussed in Section~\ref{eq:Introduction}.

\subsection{ADORA}
\label{sec:ADORA}

To better leverage the heterogeneous quality and utility of training trajectories, we propose \textbf{ADORA}, which dynamically calibrates advantage estimates by re-weighting samples according to their utility within the current epoch.
Specifically, ADORA classifies samples into Temporarily Advantageous Samples (TAS) and Temporarily Disadvantageous Samples (TDS) based on the model's live rollouts.
The core idea is to \textbf{focus the model's learning effort on TAS, with this classification evolving dynamically as training progresses}.

Formally, for each sample \( s \), we define a scalar weight \( w_s \in \mathbb{R}^+ \) and apply it to the normalized advantage:
\begin{equation}
\tilde{A}^s = w_s \cdot \hat{A}^s 
\end{equation}
\noindent where $\hat{A}^s = \{\hat{A}_{i}^s\}_{i=1}^G$ and each \(\hat{A}_i^s\) is computed according to Equation~\ref{eq:rl_adv}.

Since \( w_s \) is sample-level and independent of token-level actions, this modification preserves the unbiased nature of the policy gradient.

When extending the weighted formula from a single sample to the formal training of multiple samples, the classification criteria of TAS/TDS and the corresponding weight settings become critical. In other words, two key questions arise:
\begin{enumerate}
    \item How to determine whether a sample belongs to TAS or TDS?
    \item How to assign a corresponding weight \( w_s \) that reflects its training utility?
\end{enumerate}

\subsubsection{Criteria for Sample Differentiation}
A central challenge in RL with reasoning models is that not all successful rollouts are equally useful for driving progress.
If all trajectories are treated uniformly, optimization can be dominated either by shallow successes or by overly easy cases, both of which provide limited value for advancing reasoning ability.
ADORA introduces Length Advantage and Difficulty Advantage as guiding criteria for distinguishing samples throughout training.

\paragraph{Length Advantage.}
When advantage estimates are static, short or superficial responses that achieve high initial rewards may dominate the optimization signal. 
Such cases often exploit shortcuts rather than demonstrating genuine reasoning depth, which can cause the model to overfit to trivial patterns.
To distinguish genuine deliberation from such shortcuts, ADORA operates on a key intuition that longer successful trajectories are more likely to reflect extended deliberation, making them more valuable for cultivating robust reasoning skills.
Formally, we define a sample $s$ as having a Length Advantage if the following condition is met:
\begin{equation}
\mathrm{Len}_{\text{adv}} \iff
L_{\text{max\_succ}}^s > \bar{L}_{\text{fail}}^s
\end{equation}
\noindent where $L_{\text{max\_succ}}^s$ is the length of the longest successful rollout and $\bar{L}_{\text{fail}}^s$ is the average length of unsuccessful rollouts.

\paragraph{Difficulty Advantage.}
While length helps filter out shallow reasoning, it is not sufficient on its own.
Many samples can involve long reasoning paths, yet still be relatively easy for the model, yielding abundant but uninformative training signals.
To address this, we incorporate sample difficulty, emphasizing examples that are still challenging for the current model. 
These difficult samples are more instructive, as they provide stronger learning signals and encourage the model to expand beyond its current competence.
We consider a sample $s$ to have a Difficulty Advantage if: 
\begin{equation}
\mathrm{Diff}_{\text{adv}} \iff 
0 < R_{\text{succ}}^s \le \tau
\label{eq: diff adv}
\end{equation}
\noindent where $R_{\text{succ}}^s$ denotes the proportion of successful rollouts among all rollouts of sample $s$, and $\tau$ is a predefined threshold.

Together, Length and Difficulty Advantages offer complementary perspectives: the former filters out shallow successes, while the latter ensures that training is guided by samples that are both challenging and rich in reasoning content.

\begin{table*}[h]
\setlength{\aboverulesep}{0pt}
\setlength{\belowrulesep}{0pt}
  \caption{Avg@3 performance on various benchmarks. Dashes (–) denote unavailable official scores. \textbf{Bold} highlights the best result within each group.}
  \label{tab:pass1_accuracy}
  \centering
  \renewcommand{\arraystretch}{1.3}
  \small
  \setlength{\tabcolsep}{5pt}
  \resizebox{0.78\linewidth}{!}
{\begin{tabular}{l| c ccc c  }

  \toprule
  \multirow{2}{*}{\textbf{Model}} 
    & \multirow{2}{*}{\textbf{MathVista}} 
    & \multirow{2}{*}{\textbf{MathVerse}}
& \multirow{2}{*}{%
   \begin{tabular}[b]{@{}c@{}} 
      \textbf{MathVerse} \\[-0.3em] 
      \textbf{\scriptsize(mini\_Vision\_Only)} 
   \end{tabular}}
    & \multirow{2}{*}{\textbf{DynaMath}} 
    & \multirow{2}{*}{\textbf{Overall}} \\
    & & & & & \\
  \midrule
  Claude 3.7-Sonnet        & \textbf{66.8}  & 51.4 & 46.7 & - & - \\
  Gemini2-flash            & 59.1 & \textbf{59.3} & \textbf{47.8} & - & - \\
  \midrule
  MM-EUREKA-7B           & 72.7 & 50.6 & \textbf{48.3} & - & - \\
  MMR1-math-v0           & 70.2 & 49.8 & 45.1 & - & - \\
  Vision-R1-7B           & \textbf{73.5} & \textbf{52.4} & 46.7 & \textbf{56.3} & \textbf{57.2} \\
  \midrule
    Gemma3-4b-it & 46.3 & 25.2 & 13.5 & 10.5 & 23.88 \\
     GRPO & 47.2 & 24.9 & 13.6 & 11.0 & 24.18 \\
    \rowcolor{gray!10}
    \textbf{+ ADORA} & \textbf{48.3 (+1.1)} & \textbf{26.1 (+1.2)} & \textbf{14.5 (+0.9)} & \textbf{12.2 (+1.2)} & \textbf{25.28 (+1.1)} \\
    \midrule
    Internvl3-2b & 57.0 & 32.5 & 25.3 & 14.6 & 32.35 \\
     GRPO & 60.7 & 34.7 & 30.7 & 15.1 & 35.30 \\
    \rowcolor{gray!10}
    \textbf{+ ADORA} & \textbf{64.8 (+4.1)} & \textbf{39.2 (+4.5)} & \textbf{34.9 (+4.2)} & \textbf{18.1 (+3.0)} & \textbf{39.25 (+3.95)} \\
  \midrule
  Qwen2.5-VL-7B     & 67.3 & 46.3 & 40.2 & 50.3 & 51.0 \\
  GRPO              & 70.2 & 48.2 & 44.1 & 53.3 & 54.0 \\
  \rowcolor{gray!10}
  \textbf{+ADORA}   & \textbf{73.5 (+3.3)} & \textbf{52.9 (+4.7)} & \textbf{48.6 (+4.5)} & \textbf{58.7 (+5.4)} & \textbf{58.4 (+4.4)} \\
  \bottomrule
  \end{tabular}}
\end{table*}

\subsubsection{Adaptive Advantage for Weak and Strong Reasoning Models}
Different models exhibit distinct behaviors during RL sampling due to variations in their reasoning capabilities.
Weaker models often overfit to simple shortcuts and need guidance to develop deeper reasoning, while stronger models, already equipped with robust capabilities, benefit from strategies that emphasize challenging and instructive samples.
ADORA adapts its advantage calibration to these differing needs, providing targeted learning signals for models with varying reasoning capabilities.

Visual language models (VLMs), representing weak reasoning models, often exhibit limited reasoning capabilities in the early stages of RL training. 
During the rollout phase, responses that lack sufficient reasoning but achieve immediate rewards can dominate the optimization signal, steering the model toward shallow patterns and hindering the acquisition of advanced reasoning skills.
Consequently, ADORA employs an \textbf{attenuation} strategy, treating samples that fail to meet the Length Advantage criterion as TDS and suppressing their learning signals.
Formally, we introduce an attenuation hyperparameter $\lambda_{\text{att}} \in (0, 1)$ and define the sample weight as:
\begin{equation}
w_s =
\begin{cases}
1, & \text{if} \text{ } \mathrm{Len}_{\text{adv}} \\
\lambda_{\text{att}}, & \text{otherwise}
\end{cases}
\label{eq:att_eq}
\end{equation}
\noindent where TAS retain their full advantage signal (\( w_s = 1 \)) and TDS are down-weighted (\( w_s < 1 \)). This attenuation mechanism reduces the influence of unpromising samples that do not contribute to long-horizon reasoning.

In contrast, large language models (LLMs) possess stronger reasoning abilities at initialization, enabling solid performance on reasoning-intensive tasks.
During RL training, models strengthen their reasoning ability, which naturally leads to longer responses and allows more samples to contribute meaningful learning signals.
Accordingly, the focus shifts from denoising to breaking learning plateaus. Strong models require stronger signals from high-quality, challenging samples to continue improving. Adora therefore adopts an \textbf{amplification} strategy, identifying samples that meet both the Length and Difficulty Advantage criteria as TAS, and strengthening their contribution to the optimization process via an amplification hyperparameter $\lambda_{\text{amp}} > 1$.
We assign:
\begin{equation}
w_s =
\begin{cases}
\lambda_{\text{amp}}, & \text{if} \text{ } \mathrm{Len}_{\text{adv}} \text{ } \text{\&}\text{ }\mathrm{Diff}_{\text{adv}} \\
1,  & \text{otherwise}
\end{cases}
\label{eq:amp_eq}
\end{equation}
This amplification effect reinforces learning from challenging and instructive samples, promoting curriculum-style progression.

Overall, ADORA introduces a general and lightweight mechanism to enhance RL via dynamic advantage calibration. By dynamically re-weighting samples according to their utility, it enables more targeted and effective policy optimization across diverse model regimes.
\begin{table*}[h]
\caption{Avg@3  performance on various math benchmarks. \textbf{Bold} represents the best performance in each group.}
    \centering
    \renewcommand{\arraystretch}{1.3}
    \vspace{-0.5em}
    \small
    \resizebox{\linewidth}{!}{
    \begin{tabular}{@{}l| *{6}{c} c @{}}
    \toprule
    \multirow{1}{*}{\centering\textbf{Model}} & 
    \multicolumn{1}{c}{\textbf{GSM8K}} & \multicolumn{1}{c}{\textbf{MATH500}} & \multicolumn{1}{c}{\textbf{AMC23}} & \multicolumn{1}{c}{\textbf{CollegeMath}} & \multicolumn{1}{c}{\textbf{OlympiadBench}} & \multicolumn{1}{c}{\textbf{AIME24}} & \multicolumn{1}{c}{\multirow{1}{*}{\textbf{Overall}}} \\
    \midrule
    DeepSeek-Math-7B & 28.4 & 19.6 & 10.0 & 12.0 & 3.0 & 0.0 & 19.83 \\
     GRPO & 68.2 & 39.5 & 20.0 & 29.8 & 12.0 & 3.3 & 28.80 \\
    \rowcolor{gray!10}
    \textbf{+ ADORA} & \textbf{68.5 (+0.3)} & \textbf{41.8 (+2.3)} & \textbf{25.0 (+5.0)} & \textbf{31.6 (+1.8)} & \textbf{12.9 (+0.9)} & \textbf{3.3 (+0.0)} & \textbf{30.52 (+1.72)} \\
    \midrule
    Mistral-v0.1-7B & 21.2 & 5.4 & 0.0 & 3.8 & 2.4 & 0.0 & 5.47 \\
     GRPO & \textbf{54.0} & 26.8 & 10.0 & 11.4 & 4.1 & 0.0 & 17.72 \\
    \rowcolor{gray!10}
    \textbf{+ ADORA} & 53.8 (-0.2) & \textbf{30.4 (+3.6)} & \textbf{10.0 (+0.0)} & \textbf{12.4 (+1.0)} & \textbf{4.7 (+0.6)} & \textbf{0.0 (+0.0)} & \textbf{18.55 (+0.83)} \\
    \midrule
    Llama-3.1-8B & 40.2 & 12.7 & 2.5 & 6.4 & 3.1 & 0.0 & 10.82 \\
     GRPO & 66.1 & 33.8 & 15.0 & 22.0 & 5.3 & 0.0 & 23.72 \\
    \rowcolor{gray!10}
    \textbf{+ ADORA} & \textbf{66.7 (+0.6)} & \textbf{39.4 (+5.6)} & \textbf{15.0 (+0.0)} & \textbf{23.1 (+1.1)} & \textbf{10.5 (+5.2)} & \textbf{0.0 (+0.0)} & \textbf{25.78 (+2.06)} \\
    \midrule
    Qwen2.5-7B & 56.3 & 57.2 & 37.5 & 24.3 & 26.3 & 10.0 & 35.27 \\
    GRPO & 89.1 & 73.2 & 50.0 & 28.6 & 35.1 & 13.3 & 48.22 \\
    \rowcolor{gray!10}
    \textbf{+ADORA} & \textbf{89.6 (+0.5)} & \textbf{76.2 (+3.0)} & \textbf{62.5 (+12.5)} & \textbf{29.3 (+0.7)} & \textbf{36.0 (+0.9)} & \textbf{16.7 (+3.4)} & \textbf{51.72 (+3.50)} \\
    \bottomrule
    \end{tabular}
    }

    \label{tab:distill_vs_rl}
\end{table*}
\section{Experiment}

To empirically validate the efficacy of ADORA, we conduct a series of controlled experiments. Section \ref{experiment: VLM} first reports the results of VLM geometry reasoning tasks and Section \ref{experiment: LLM} presents the results of LLM mathematical reasoning tasks.

\paragraph{Setup.} 
We employ GRPO as the base RL algorithm, with ADORA integrated upon it. All RL experiments are implemented under the verl framework \cite{sheng2024hybridflow}, utilizing Math-Verify for rule-based outcome verification.
For VLM tasks, all experiments are conducted using 2,000 samples from the Geometry3K training set \cite{lu2021inter}.
For LLM tasks, we conduct RL training on using the MATH500 training set \cite{lightman2023let}, which contains 12,000 samples. 
In our experiments, we set the hyperparameters as follows: the threshold $\tau=0.5$, the attenuation weight $\lambda_{\text{att}}=0.1$, the amplification weight $\lambda_{\text{amp}}=2$, and additional detailed training hyperparameter settings are provided in Appendix~\ref{sec:hyperparameters}.
To ensure reproducibility and robustness, we conduct three separate runs and report the average performance to mitigate random variations. 

\paragraph{Evaluation.} For evaluation, VLM performance is primarily assessed on MathVista \cite{lu2023mathvista}, Math Verse \cite{zhang2024mathverse} and DynaMath \cite{zou2024dynamathdynamicvisualbenchmark} datasets. 

For evaluation on LLM tasks, we mainly focus on seven widely used math reasoning benchmarks, including GSM8K \cite{cobbe2021training}, MATH500 \cite{hendrycks2021measuring}, AMC23 \cite{li2024numinamath}, CollegeMath \cite{tang2024mathscale}, OlympiadBench \cite{he2024olympiadbench}, and AIME24. 
For all these benchmarks, we report the avg@3 results and setting the sampling temperature to 0.

\subsection{VLM}\label{experiment: VLM}

\paragraph{Baselines.} Recent works\cite{meng2025mm,MMR1-Math2025,huang2025vision} have reproduced R1 on VLMs.
We take these methods as baselines for comparison and further analyze the amount of training data consumed by different approaches (see Appendix~\ref{sec:dataresource}). 
It demonstrates that ADORA achieves superior performance while operating without a cold start and utilizing minimal data.
Furthermore, we conduct RL experiments across multiple model families—including Qwen \cite{bai2025qwen2}, Gemma \cite{team2025gemma}, and InternVL \cite{zhu2025internvl3}—evaluating ADORA against the vanilla GRPO baseline.

\paragraph{Results.} The results in Table~\ref{tab:pass1_accuracy} indicate that ADORA consistently outperforms the vanilla GRPO baseline across diverse model families and varying parameter scales. Compared to other open-source models with 7B parameters, ADORA delivers the best performance. On MathVista, it matches Vision-R1-7B~\cite{huang2025vision} and substantially outperforms advanced closed-source models. Moreover, on MathVerse and DynaMath, ADORA surpasses peer models of the same size by a large margin.
In conjunction with Table \ref{tab:data_comparison_transposed_version}, ADORA does not rely on the cold-start and achieves state-of-the-art (SOTA) performance on nearly all benchmarks with only 2,000 samples. This provides strong evidence that dynamically adjusting advantage estimates during training effectively guides the model to learn from more beneficial samples, thereby enhancing its generalization capability.

\subsection{LLM}\label{experiment: LLM}

\paragraph{Baselines.}
Employing models from the Qwen \cite {yang2024qwen2}, Mistral, and LLaMA \cite {grattafiori2024llama}families, we compare ADORA against the vanilla GRPO baseline to evaluate its effectiveness.

\paragraph{Results.} 
As shown in Table \ref{tab:distill_vs_rl}, training with ADORA consistently improves the performance of vanilla GRPO across a range of mathematical reasoning benchmarks.
Specifically, ADORA boosts the overall average performance across all model families by margins ranging from 0.83\% to 3.50\%.
Notably, substantial gains are achieved on challenging datasets such as AMC23 and MATH500, while improvements are consistently observed across all remaining tasks.
In summary, these results confirm that ADORA is a robust, versatile, and effective plug-and-play enhancement. It is universally applicable across diverse model architectures, yielding the most significant improvements on tasks demanding complex reasoning when integrated with GRPO.

\section{Analysis}
Beyond achieving superior aggregate performance, an understanding of how ADORA improves reasoning is crucial. 
This section analyzes ADORA's impacts on model behavior and learning characteristics.
First, Section~\ref{training_compare} compares ADORA and vanilla GRPO throughout the training process and analyzes the notable thinking patterns evolved by the model.
Subsequently, we conduct a comprehensive series of ablation studies in Section~\ref{sec:ablation} to verify the method's robustness and design choices, covering hyperparameter sensitivity, the formulation of ADORA's advantage criteria, and the effectiveness of integrating with different RL algorithms.

\subsection{Empirical Study}
\label{training_compare}

\begin{figure}[t]
    \centering
    \begin{subfigure}{0.48\linewidth}
        \centering
        \includegraphics[width=\linewidth]{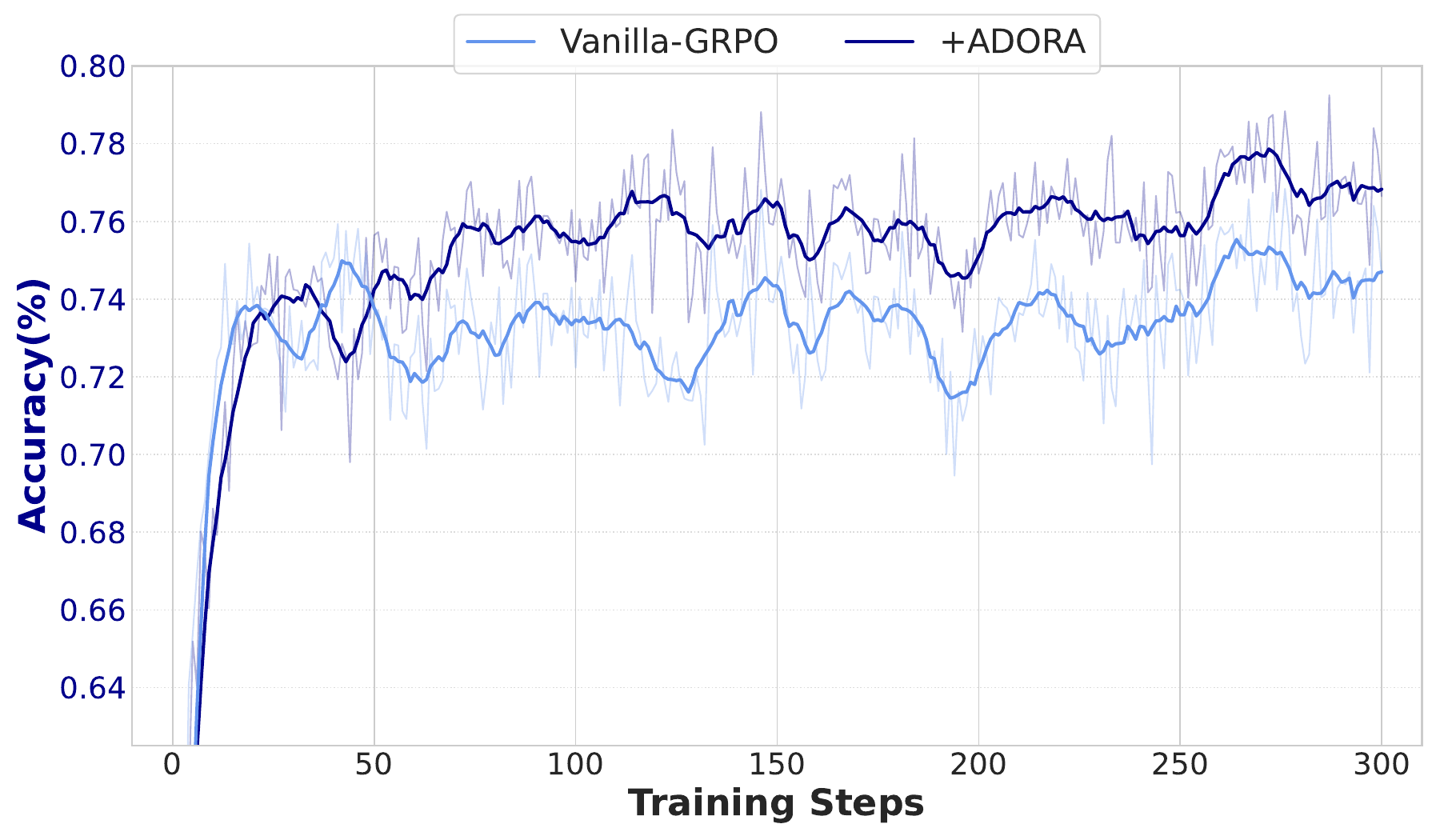}
        \caption{LLM}
        \label{fig:math_performance}
    \end{subfigure}
    \hfill %
    \begin{subfigure}{0.48\linewidth}
        \centering
        \includegraphics[width=\linewidth]{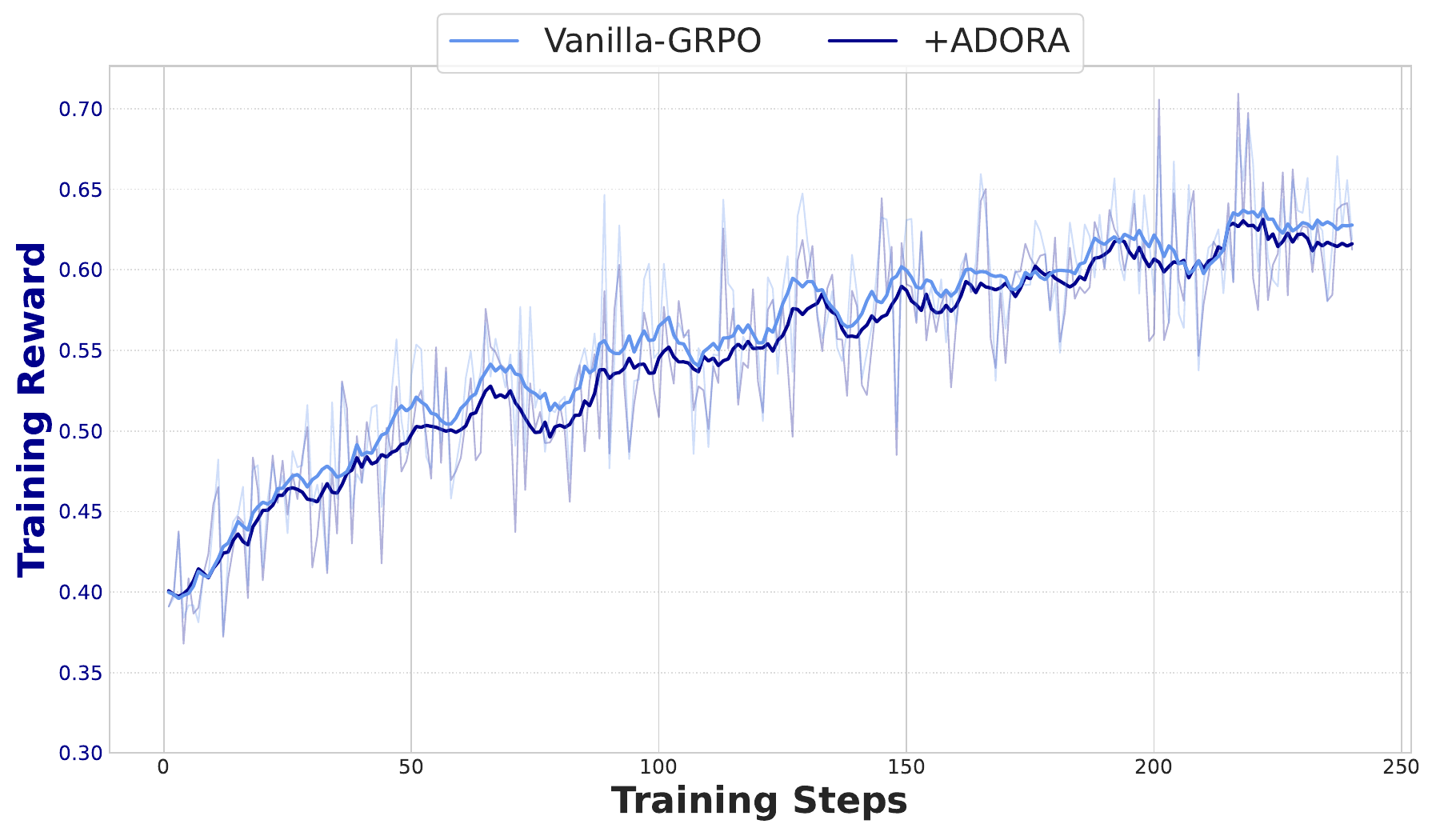}
        \caption{VLM}
        \label{fig:geo_performance}
    \end{subfigure}
    
    \caption{Comparison of vanilla GRPO vs. integration with ADORA for the training of Qwen models.}
    \label{fig:main_figure}
    \vspace{-1em}
\end{figure}

\paragraph{Training Comparison.} 
Figure~\ref{fig:math_performance} compares ADORA with the GRPO baseline throughout training on LLM tasks.
While the performance gap is modest in early stages, a clear divergence emerges as training progresses. Uniform weighting in vanilla GRPO induces diminishing returns, where redundant or noisy samples (TDS) hinder further improvement. 
Conversely, ADORA selectively amplifies the reward signal of high-value samples (TAS), increasing the marginal effectiveness of each update. 
This results in two key advantages:
(i) \emph{Superior efficiency}—ADORA reaches a reward of 0.75 within 100 steps, whereas GRPO fails to do so even after 250 steps;
(ii) \emph{Higher performance ceiling}—by dynamically suppressing noise and prioritizing reasoning-intensive and difficult samples, ADORA achieves stronger performance under the same data budget.

In contrast to the LLM setting, Figure~\ref{fig:geo_performance} reveals a distinct training trajectory for VLMs. 
ADORA's attenuation mechanism yields smaller gradient updates on TDS, leading to training performance comparable to or slightly below Vanilla-GRPO.
However, while the baseline may achieve a marginally higher training reward by over-fitting to visual shortcuts or low-quality reasoning patterns inherent in the training set, ADORA effectively suppresses these noisy signals. 
By prioritizing samples with a genuine Length Advantage, the model is compelled to develop more robust and intrinsic reasoning capabilities rather than relying on superficial correlations. 
This advantage is clearly evidenced in the out-of-domain downstream tasks presented in Table~\ref{tab:pass1_accuracy}, where ADORA-trained models consistently outperform the baseline.

\begin{figure}[t]
    \centering
    \vspace{-1em}
    \includegraphics[width=1\linewidth]{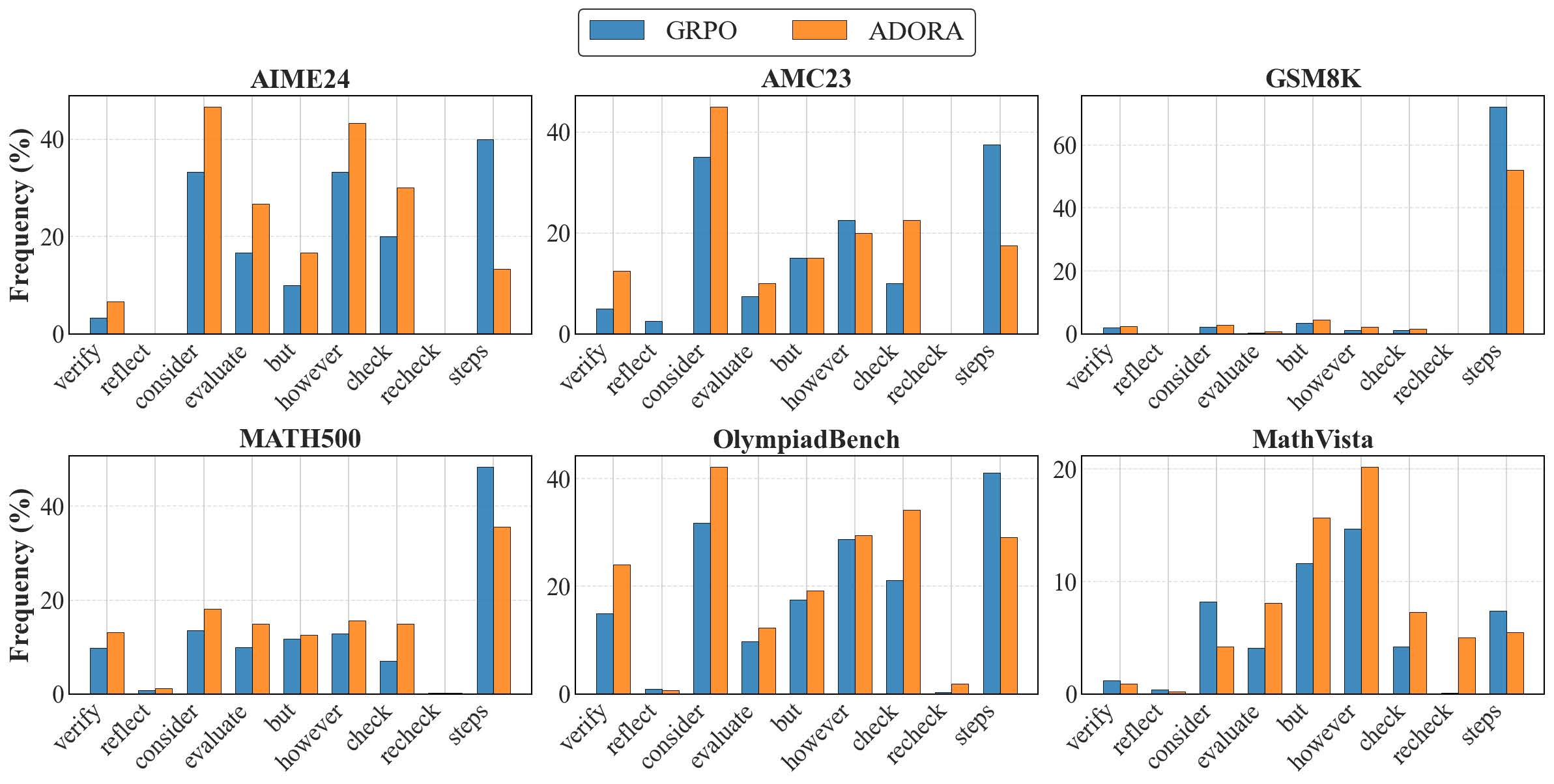}
    \caption{Distribution of Reasoning-Related Keywords for ADORA and vanilla GRPO.}
    \label{fig:rl}
    \vspace{-1em}
\end{figure}

\paragraph{Thinking Pattern.} 
To analyze how ADORA reshapes the model beyond final accuracy, we investigate the reasoning behaviors of models trained with ADORA and vanilla GRPO.

One of the most direct indicators of explicit reasoning is the frequency of reflective vocabulary.
As shown in Figure~\ref{fig:rl}, ADORA-trained models exhibit two prominent linguistic trends: increased use of core reflective terms (e.g., \emph{verify}, \emph{evaluate}, \emph{consider}, \emph{check}) and more frequent transitional markers (e.g., \emph{but}, \emph{however}), both signaling structured and deliberate reasoning.
In contrast,  the frequency of the word \emph{step}—which often signifies a rigid, formulaic thinking mode—drops significantly.
This shift indicates that ADORA encourages models to prioritize self-monitoring and logical verification, facilitating a transition from rigid imitation toward more autonomous and reflective reasoning.

The right-shifted and heavier-tailed token length distributions in Figure~\ref{fig:generation_length} show that ADORA produces longer answers across benchmarks, reflecting its distinct thinking patterns compared to vanilla GRPO.
To distinguish whether this increase reflects deeper reasoning on difficult problems or unnecessary overthinking, we introduce the \emph{Overthinking Score}~\cite{cuadron2025danger}. A higher score indicates a greater degree of overthinking.
As shown in Table~\ref{tab: overthinking}, ADORA demonstrates an adaptive calibration of reasoning depth: on the simpler GSM8K, it maintains a score comparable to GRPO, whereas on the challenging AIME24, it achieves a significantly lower overthinking score (40.1 vs. 44.8). This suggests that ADORA effectively distinguishes between productive reflection and redundant computation, encouraging deeper reasoning only when task complexity demands it.

\begin{figure}[t]
    \centering
    \includegraphics[width=1\linewidth]{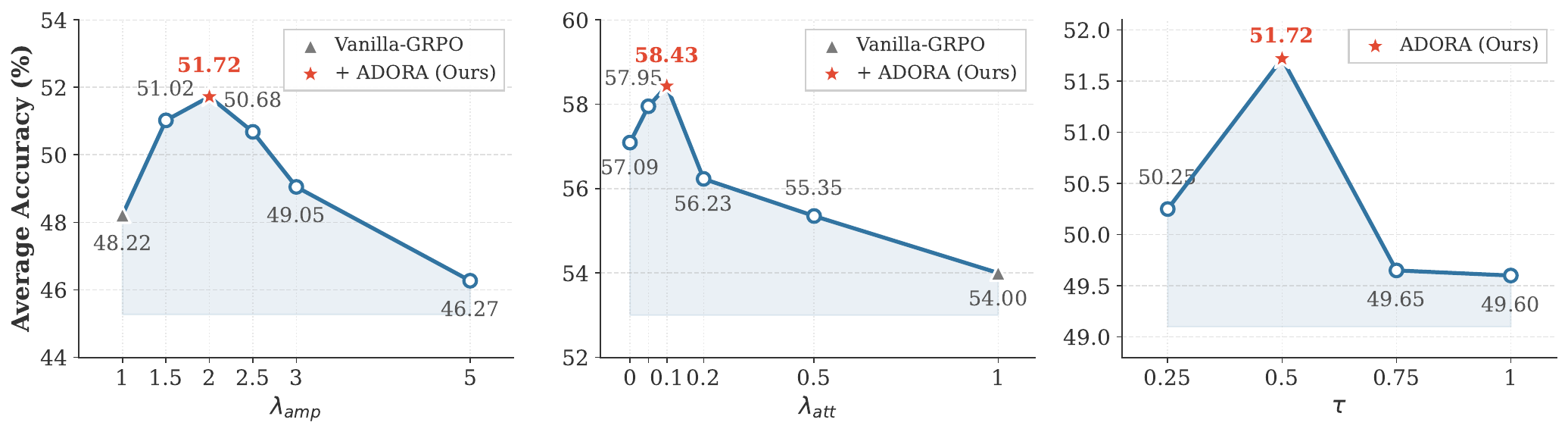}
        \vspace{-2em}
    \caption{Hyperparameter ablation of $\tau$, $\lambda_{\text{att}}$ and $\lambda_{\text{amp}}$}
    \label{fig:rl}
    \vspace{-1em}
\end{figure}

\subsection{Ablation Study}
\label{sec:ablation}

\paragraph{Hyperparameter Sensitivity.} We perform an ablation experiment on the difficulty threshold in Eq.~\ref{eq: diff adv}. Following standard principles from curriculum learning and adaptive sampling, a sample’s priority is increased whenever its accuracy falls below this threshold, reflecting the need to allocate more learning resources to harder problems. We ablate the threshold over $\tau \in \{0.25, 0.5, 0.75, 1\}$ and find that $\tau = 0.5$ consistently achieves the best balance between easy tasks (GSM8K) and hard tasks (AIME24), effectively \emph{distinguishing between "mastered" and "unmastered" samples.}

For the attenuation weight in Eq.~\ref{eq:att_eq}, our experiments show that when $\lambda_{\text{att}} < 1$, TDS samples are effectively down-weighted, consistently outperforming vanilla GRPO. The best results are obtained within $\lambda_{\text{att}} \in [0.05, 0.2]$, confirming that for VLMs with relatively weaker reasoning abilities, attenuation reliably acts as a denoising mechanism, preventing the policy from being misled by low-quality rollouts.

For LLM training, Performance consistently exceeds the vanilla-GRPO for amplification weights in the range $\lambda_{\text{amp}} \in (1, 3]$, with a stable optimum observed around $\lambda_{\text{amp}} \in [1.5, 2.5]$. This indicates that for models with strong reasoning capabilities, amplifying the gradients of high-quality and difficult samples provides a training benefit.

\paragraph{Advantage Criteria.}
We investigate the impact of different advantage criteria in ADORA by comparing the effects of using Length Advantage, Difficulty Advantage, and their combination on LLM and VLM tasks. 
As shown in Figure~\ref{fig:llm radar}, for LLM tasks, the joint criterion consistently outperforms either individual criterion across multiple benchmarks. Among the single criteria, the Length Advantage is more effective on moderately difficult tasks, while the Difficulty Advantage yields clear benefits on harder benchmarks.
We further evaluate the performance of VLMs using the joint criterion (in Figure~\ref{fig:vlm radar}), yet observe no significant gains over applying Length Advantage alone. This suggests that the primary benefit for VLMs stems from filtering out shallow or spurious reasoning patterns—essentially a denoising process facilitated by Length Advantage—rather than from the explicit prioritization of task complexity.

\begin{figure}[t]
\includegraphics[width=1\linewidth]{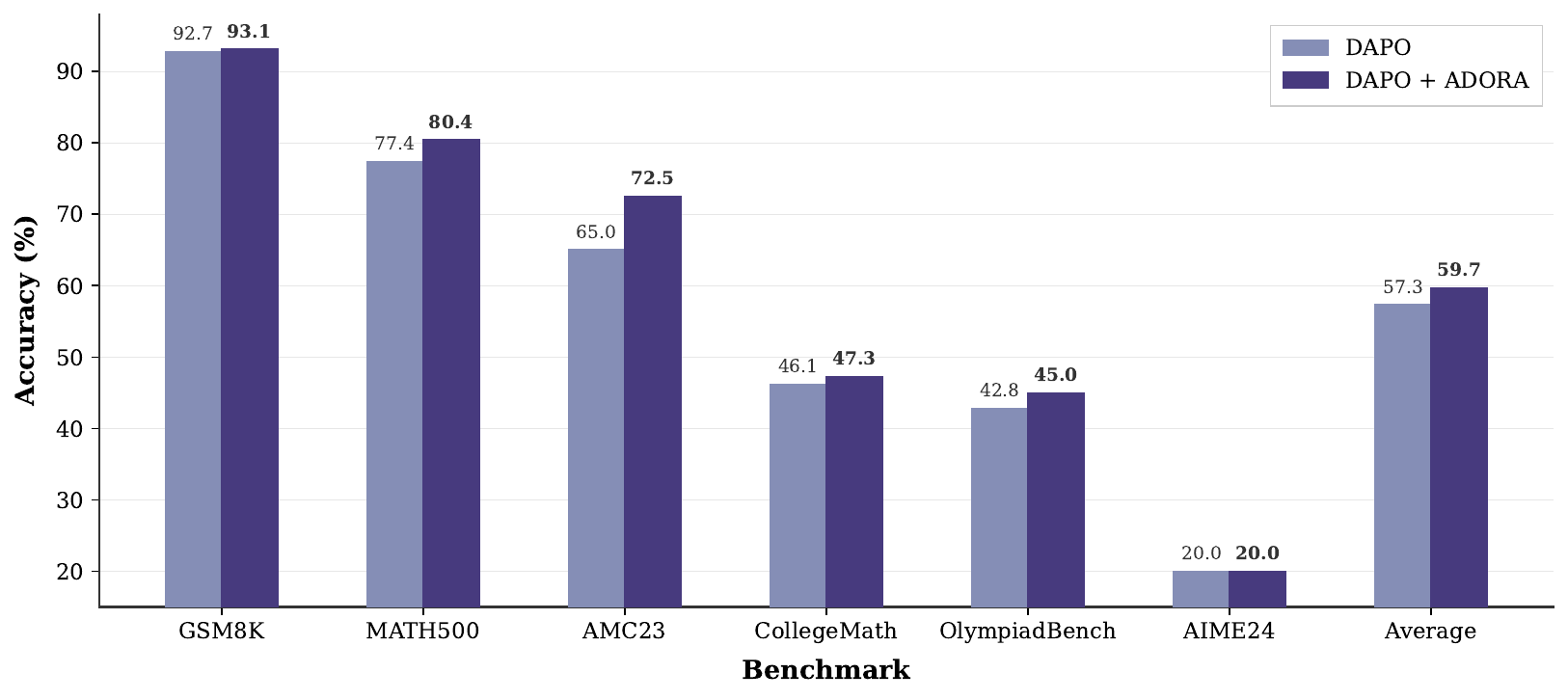}
        \vspace{-0.5em}
    \caption{Comparison between DAPO baseline and ADORA.}
    \label{fig:dapo}
    \vspace{-1em}
\end{figure}

\paragraph{RL Algorithms.}
To verify the generality of ADORA across different RL training algorithms, we additionally conduct ablation experiments by integrating ADORA with DAPO~\cite{yu2025dapo}. ADORA further enhances the already strong DAPO baseline, increasing its overall accuracy from 57.5\% to 59.9\% (see Figure~\ref{fig:dapo}). These results demonstrate that ADORA consistently improves performance across different RL algorithms, confirming its broad applicability.

\section{Conclusion}
ADORA dynamically calibrates reinforcement learning advantages via online rollouts, significantly enhancing reasoning performance and efficiency for both LLMs and VLMs by differentiating sample utility. Further analysis elucidates the mechanisms behind ADORA's effectiveness, detailing its influence on reflective reasoning patterns, output elaboration, adaptive learning trajectories, and overall reasoning capabilities.

\section*{Limitations}
While ADORA demonstrates consistent improvements, several limitations remain. First, the task-specific differentiation strategies may require redesign when applied to new domains, limiting out-of-the-box generalizability. And ADORA's efficacy is tied to rollout quality; if the base model produces low-quality reasoning trajectories, the TAS/TDS classification may become unreliable. What's more, our evaluation primarily focuses on mathematical and geometric reasoning, leaving applicability to other scenarios (e.g., commonsense reasoning, agentic tasks) unexplored. Finally, integration with advanced RL techniques beyond GRPO warrants further investigation.

\bibliography{custom}

\clearpage
\appendix

\section{Training Details}
\subsection{Training Hyperparameters}
\label{sec:hyperparameters}
The detailed training hyperparameters are provided in Tables \ref{tab:vlm_setting} and \ref{tab:llm_setting}, and all experiments are conducted on 8 NVIDIA A100 GPUs, each equipped with 80 GB of memory.

\begin{table}[h]
\small
\centering
\caption{Key hyperparameters for VLM training.}
\label{tab:vlm_setting}
\begin{tabular}{@{}c|c@{}}
\toprule
\textbf{Name} & \textbf{Value} \\ \midrule
Rollout num & 8 \\
Train batch size & 128 \\
Rollout temperature 1.0 \\
Mini batch size & 128 \\
Micro batch size per GPU & 2 \\
Learning rate & 1.0e-6 \\
Entropy coefficient & 0.0 \\
KL loss coefficient & 0.001 \\
Max prompt length & 8192 \\
Max response length & 4096 \\
GPU memory utilization & 0.7 \\
\bottomrule
\end{tabular}
\end{table}

\begin{table}[h]
\small
\centering
\caption{Key hyperparameters for LLM training.}
\label{tab:llm_setting}
\begin{tabular}{@{}c|c@{}}
\toprule
\textbf{Name} & \textbf{Value} \\ \midrule
Rollout num & 8 \\
Train batch size & 256 \\
Rollout temperature 1.0 \\
Mini batch size & 128 \\
Micro batch size per GPU & 2 \\
Learning rate & 1.0e-6 \\
Entropy coefficient & 0.0 \\
KL loss coefficient & 0.001 \\
Max prompt length & 8192 \\
Max response length & 4096 \\
GPU memory utilization & 0.7 \\
\bottomrule
\end{tabular}
\end{table}

\subsection{Comparison of Dataset Sizes}
\label{sec:dataresource}
Table~\ref{tab:data_comparison_transposed_version} summarizes the training resource configurations of ADORA and other baselines, detailing the amount of data consumed at different post-training stages. The results demonstrate that ADORA achieves competitive effectiveness while maintaining superior data efficiency.
\begin{table}[h]
    \footnotesize
    \caption{Cold-Start and RL training data comparison of multimodal methods.} 
    \centering
    \setlength{\tabcolsep}{3pt}
    \renewcommand{\arraystretch}{1.3} 
    \begin{tabular}{@{} l l l@{}} 
    \toprule
    
    \textbf{Model} & \textbf{Cold-Start Data} & \textbf{RL Data} \\ 
    \midrule
    
    MM-EUREKA-7B & 54k (open-source) & 9.3k (open-source) \\
    MMR1-math-v0 & None & 6k (open-source) \\
    Vision-R1-7B & 200k (synthetic data) & 10k (open-source) \\
    \textbf{ADORA (ours)} & None & 2k (open-source) \\
    \bottomrule
    \end{tabular}%
    \label{tab:data_comparison_transposed_version}
\end{table}

\section{Additional Experiments}

\subsection{Data Scalability}
A key question is whether ADORA's benefits persist as training data scales. As shown in Table~\ref{tab:qwen_multimodal_math}, We scaled the VLM training set from 2k to 10k samples. Results indicate that ADORA maintains a robust lead over GRPO. Crucially, ADORA trained on 2k samples (73.5\%) outperforms GRPO trained on 10k samples (71.6\%), highlighting its extreme sample efficiency. With 10k samples, ADORA further improves to 74.4\%. This trend shows that ADORA continues to amplify the marginal benefits of sample selection as data volume increases, making it increasingly effective in larger-scale settings.

\begin{figure}[h]
\includegraphics[width=1\linewidth]{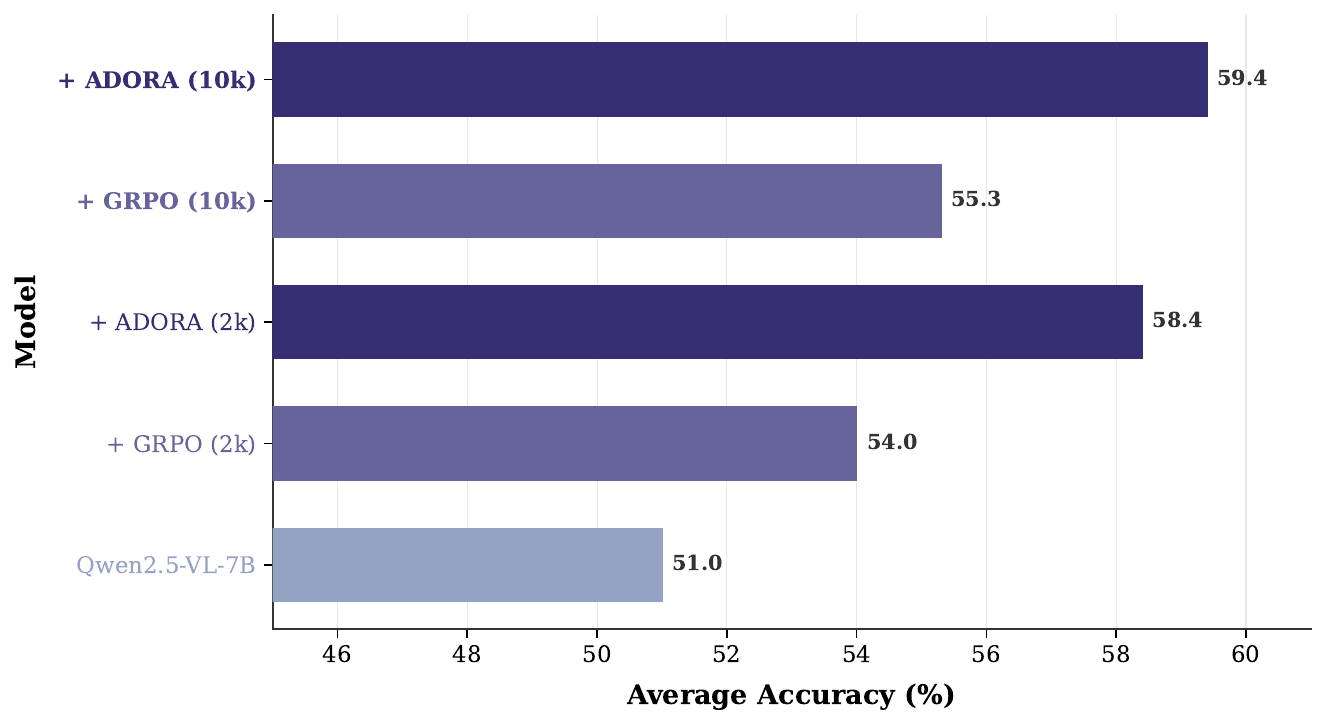}
        \vspace{-0.5em}
    \caption{Comparison between training with 2K samples and 10K samples.}
    \label{fig:210K}
    \vspace{-1em}
\end{figure}

\begin{table}[h]
\caption{Zero-shot Avg@3 performance on various multimodal math benchmarks based on Qwen2.5-VL-7B. \textbf{Bold} denotes the best performance within each training step group.}
    \centering
    \renewcommand{\arraystretch}{1.3}
    \vspace{-0.5em}
    \small
    \resizebox{\linewidth}{!}{
    \begin{tabular}{@{}l| *{4}{c} c @{}}
    \toprule
    \multirow{1}{*}{\centering\textbf{Model}} & 
    \multicolumn{1}{c}{\textbf{MathVista}} & 
    \multicolumn{1}{c}{\textbf{MathVerse}} & 
    \multicolumn{1}{c}{\shortstack{\textbf{MathVerse} \\ \textbf{\scriptsize(mini\_Vision\_Only)}}} & 
    \multicolumn{1}{c}{\textbf{DynaMath}} & 
    \multicolumn{1}{c}{\textbf{Avg.}} \\
    \midrule
    Qwen2.5-VL-7B & 67.3 & 46.3 & 40.2 & 50.3 & 51.0 \\
    \midrule
    + GRPO (2k) & 70.2 & 48.2 & 44.1 & 53.3 & 54.0 \\
    \rowcolor{gray!10}
    \textbf{+ ADORA (2k)} & \textbf{73.5} & \textbf{52.9} & \textbf{48.6} & \textbf{58.7} & \textbf{58.4} \\
    \midrule
    + GRPO (10k) & 71.6 & 50.6 & 45.3 & 53.8 & 55.3 \\
    \rowcolor{gray!10}
    \textbf{+ ADORA (10k)} & \textbf{74.4} & \textbf{53.5} & \textbf{50.1} & \textbf{59.8} & \textbf{59.4} \\
    \bottomrule
    \end{tabular}
    }
    \label{tab:qwen_multimodal_math}
\end{table}

\subsection{How Adora affects the learning trajectory of RL?} 
\label{learning_trajectory}

Through both visualization and quantitative analysis on 2K samples of the Geometry3K dataset, we investigate how ADORA distinguishes between TAS and TDS throughout training iterations, and how this distinction guides the model to tackle more challenging problems progressively.

\begin{figure}[!h]
\centering
\includegraphics[width=1\linewidth]{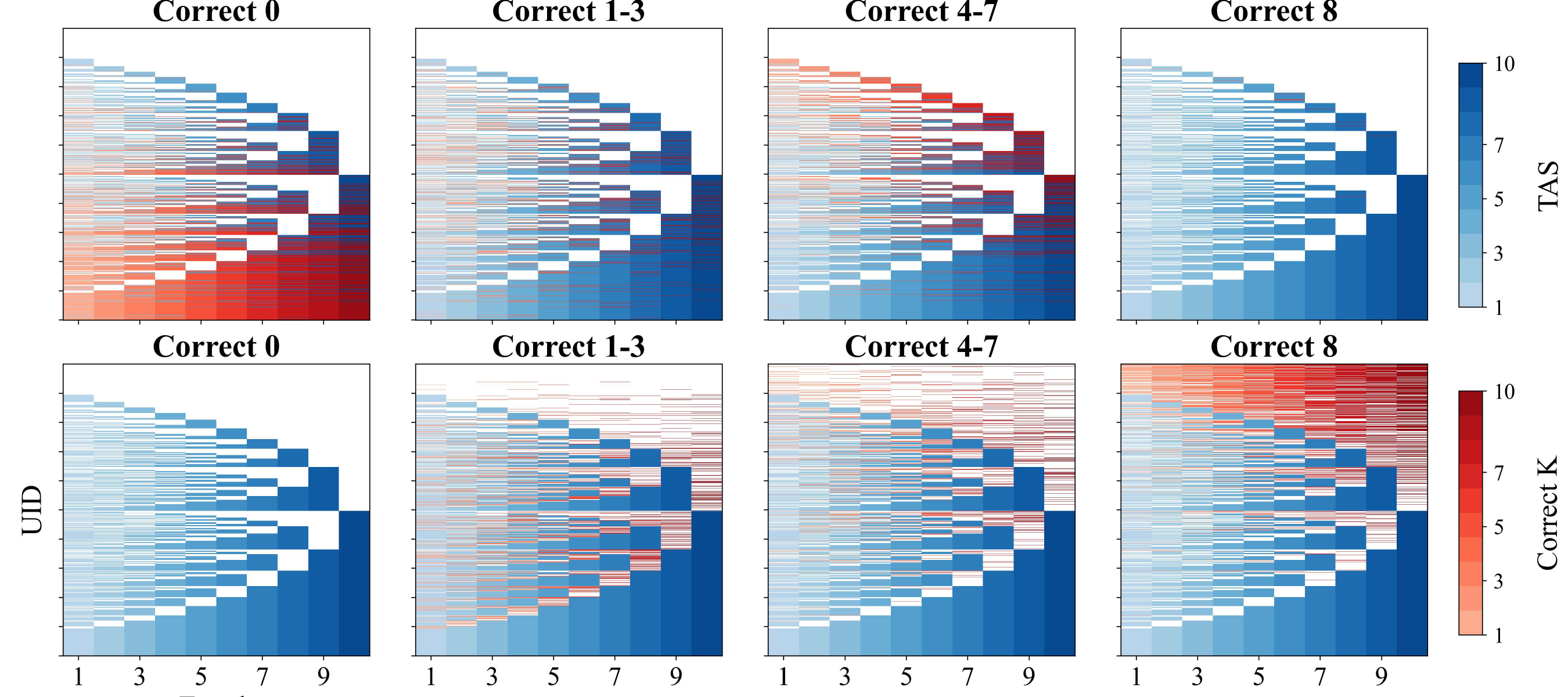}
\caption{The blue sections represent the samples selected for each epoch (clustered for easier visualization), while the red sections illustrate the distribution of samples under different Correct N settings in one sampling, representing the difficulty of the samples, both of which gradually deepen as epochs progress. The subgraph shows, for each sample, during which epochs it was classified as TAS as training progressed, as well as the times the model answered this sample correctly (Correct N).}
\label{fig:uid_summary}
\end{figure}

\begin{figure}[h]
    \centering
    \includegraphics[width=1\linewidth]{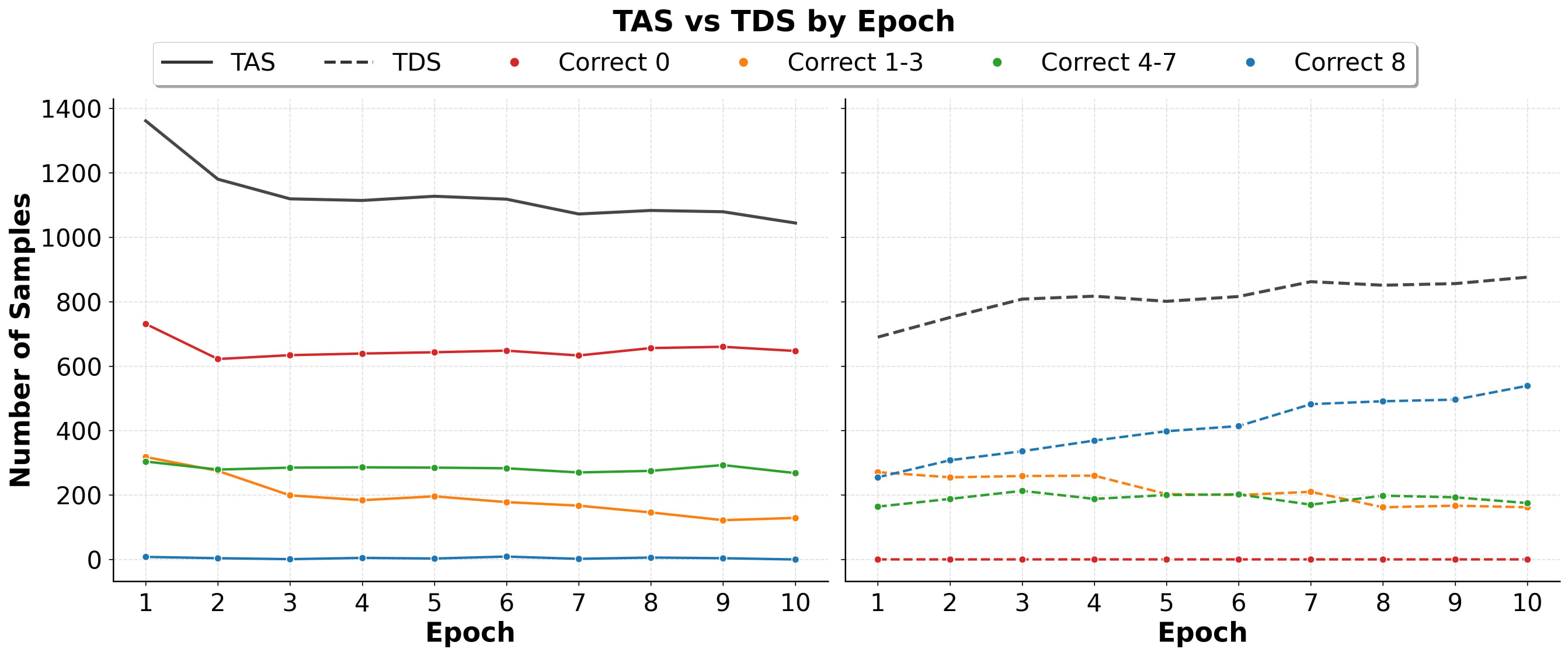}
    \caption{The changes in the number of samples of each difficulty level for the two corresponding categories of samples across epochs.}
    \label{fig:combined_uids}\
    \vspace{-1em}
\end{figure}

Figure~\ref{fig:uid_summary} and Figure~\ref{fig:combined_uids} reveal that ADORA performs better when selecting half of the data in each epoch, and the number of ``selected samples" decreases as the epochs progress.
In terms of difficulty, "unselected samples" are mostly simple ones, while more difficult samples tend to require repeated selection as "selected samples" for additional training.
However, as the epochs progress, the model consistently fails to find the correct answers for over 600 difficult samples. Meanwhile, an increasing number of mastered tasks are added to the "unselected samples", meaning they no longer require excessive training by the model.

Compared to the vanilla GRPO method, ADORA employs an ``Easy to hard; iterate if challenged" optimization strategy in its learning trajectory, enabling the model to build a more robust capability reserve when tackling subsequently harder samples. This dynamic sample prioritization mechanism not only accelerates the model’s generalization on medium‑difficulty examples but also significantly reduces redundant training on easy ones, making it a key factor in ADORA’s performance breakthroughs on geometry reasoning tasks.

\subsection{PASS@K: ADORA vs. GRPO}
\label{PASS@K}

\begin{figure}[h]
    \centering
    \vspace{-0.5em}
    \includegraphics[width=1\linewidth]{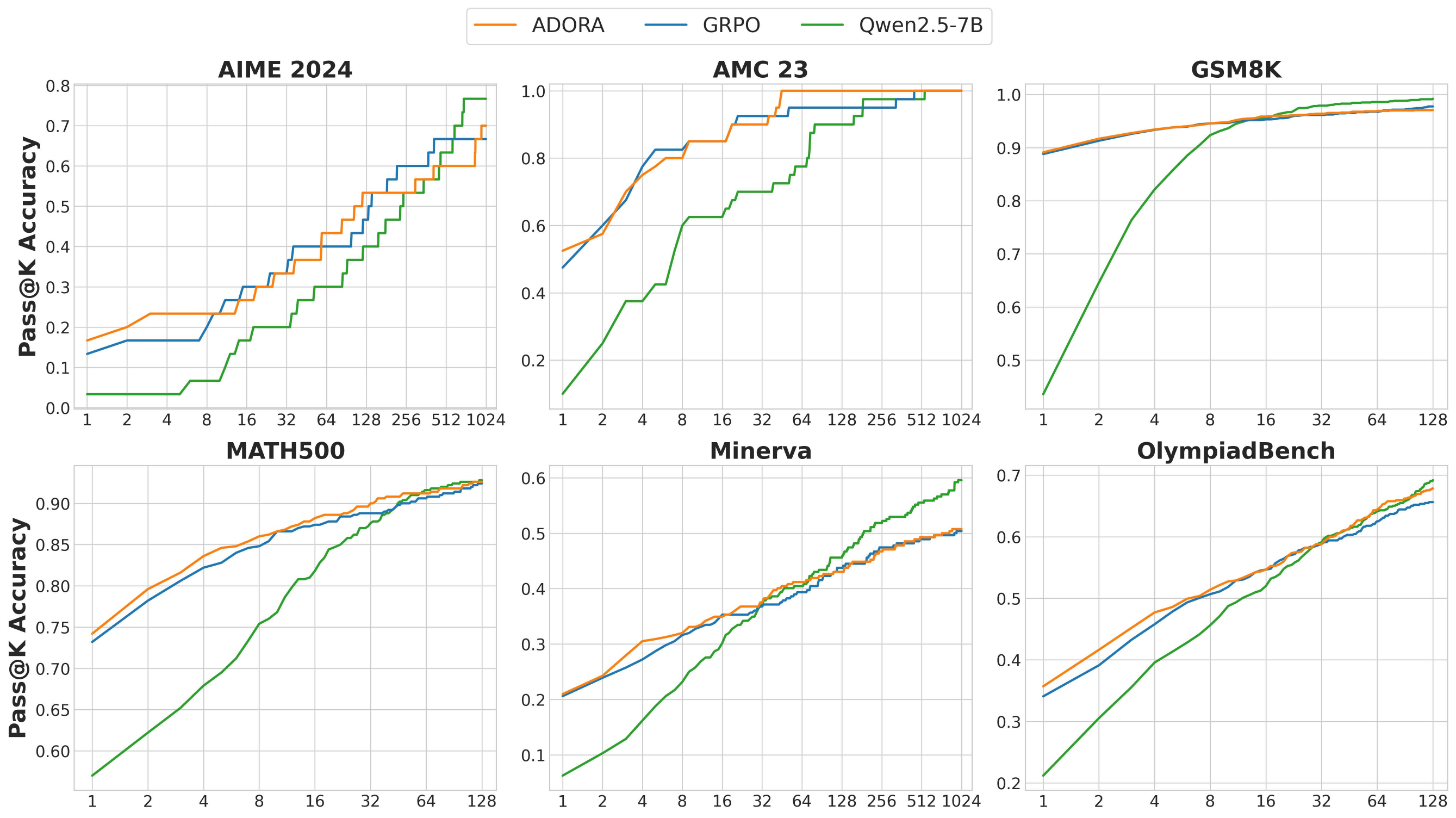}
    \caption{Pass@k curves of base model and ADORA/GRPO across multiple mathematical benchmarks.
}
    \label{fig:pass_k}
    \vspace{-1em}
\end{figure}
The Pass@K metric, which assesses if a model can correctly solve a problem in at least one of K attempts (thus indicating its upper-bound reasoning capability), was used to compare ADORA against GRPO in Figure~\ref{fig:pass_k}. 
Consistent with prior findings \cite{yue2025does}, We manually inspect to ensure that the problem-solving process is not coincidental and observe that ADORA consistently outperformed or matched GRPO across benchmarks, with both RL methods significantly surpassing the base model at smaller K values. Interestingly, while the base model sometimes overtook both at larger K, ADORA notably achieved 100\% accuracy on the AMC dataset with fewer than 64 samples, outperforming both GRPO and the base model.

These Pass@K comparisons highlight ADORA's strength: it not only improves efficiency in reaching known solutions but also appears to expand the set of viable reasoning paths the model can explore. This creates a broader "solvable problem space," enabling ADORA-trained models, given enough attempts, to solve problems where GRPO-trained counterparts might still struggle.

\subsection{Thinking Pattern}
ADORA exhibits a moderate and controlled increase in response length compared to GRPO, ensuring the generation remains efficient.
\begin{figure}[h]
    \centering
    \includegraphics[width=1\linewidth]{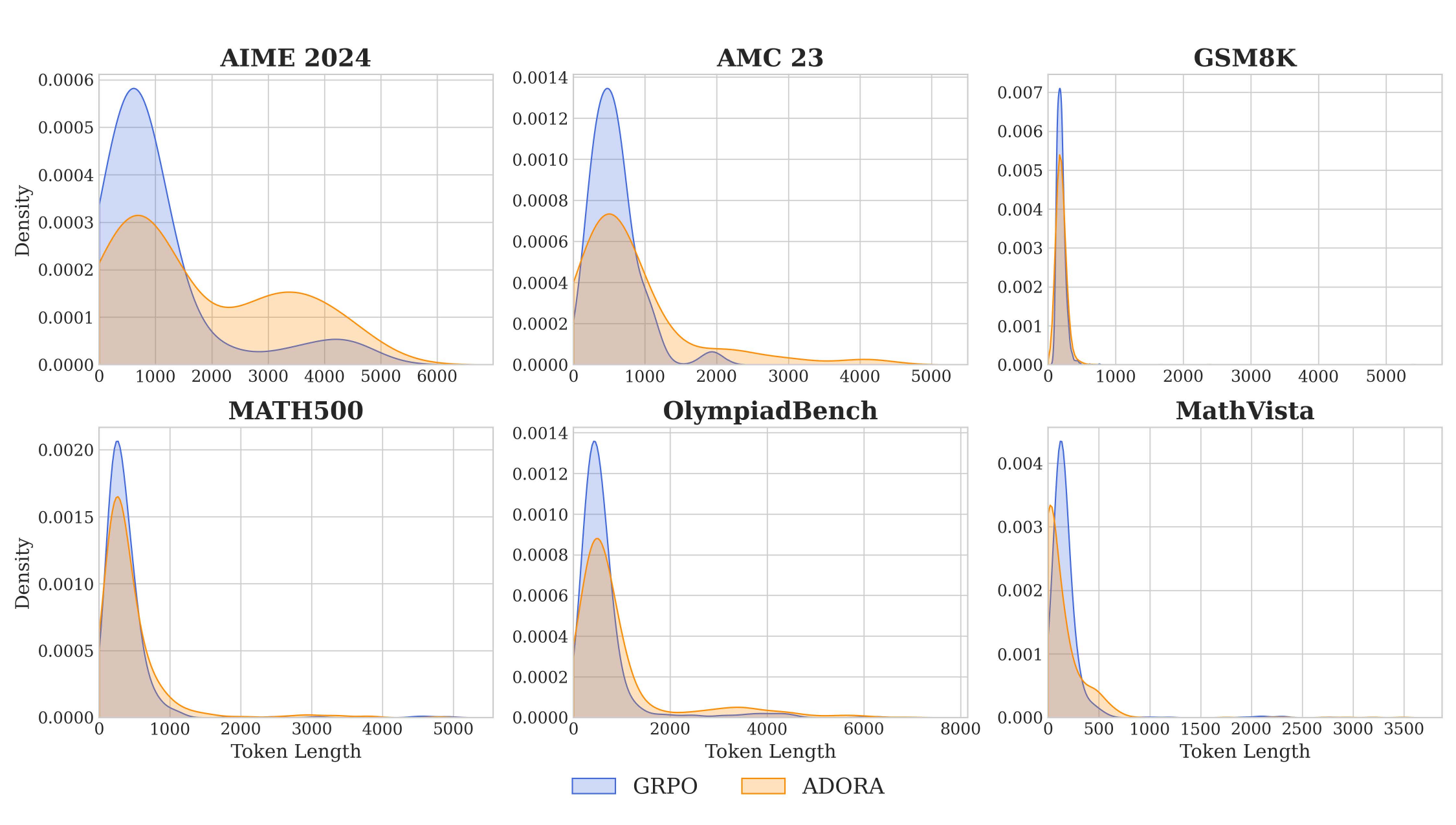}
    \caption{Comparison of Token Length Distributions Generated by GRPO and ADORA across Various Reasoning Benchmarks.
}
\vspace{-1em}
\label{fig:generation_length}
\end{figure}

\subsection{Results of Advantage Criteria Ablation}

\begin{figure}[h]
    \centering
    \begin{subfigure}{0.48\linewidth}
        \centering
        \includegraphics[width=\linewidth]{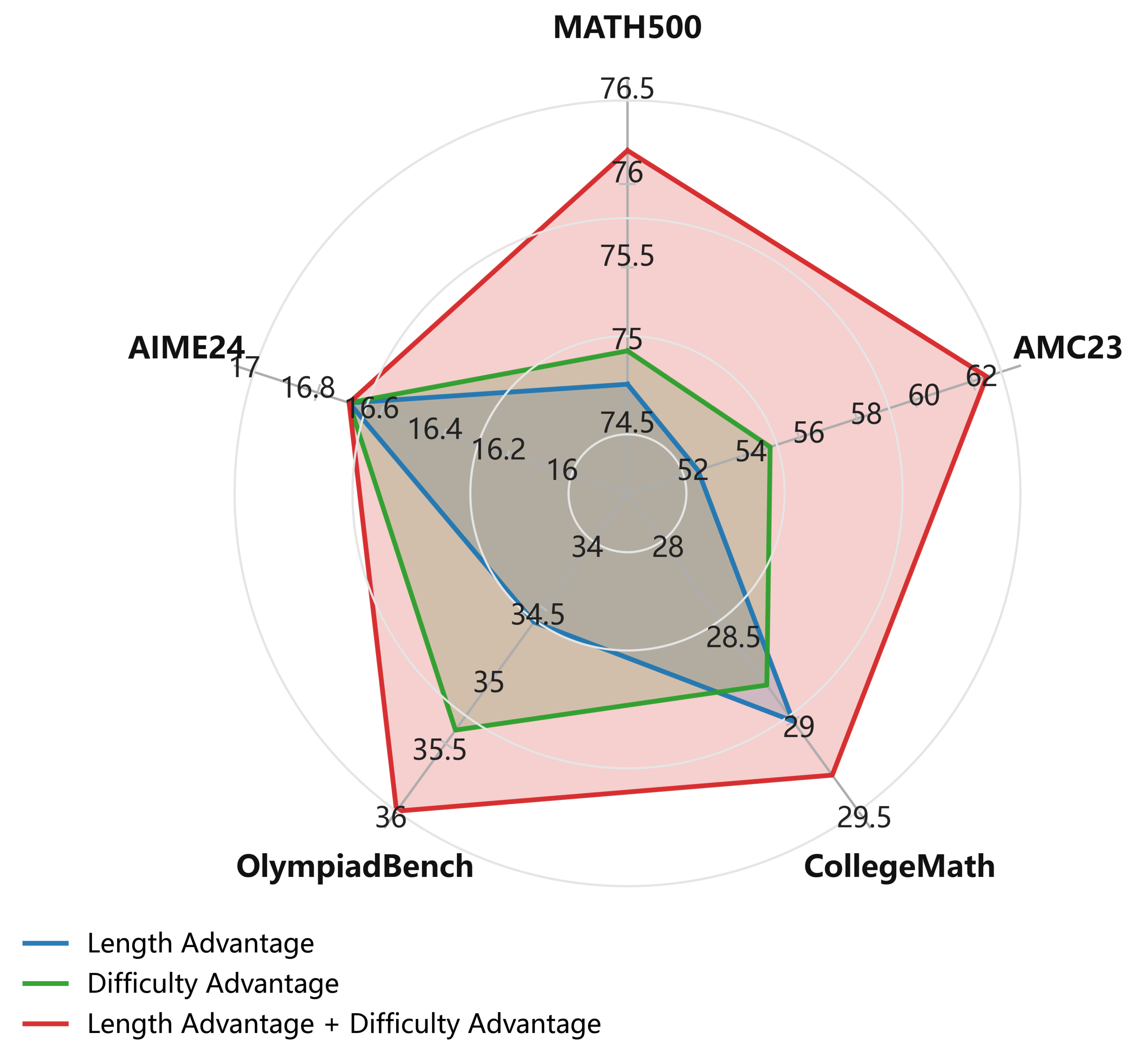}
        \caption{LLM}
        \label{fig:llm radar}
    \end{subfigure}
    \hfill
    \begin{subfigure}{0.43\linewidth}
        \centering
        \includegraphics[width=\linewidth]{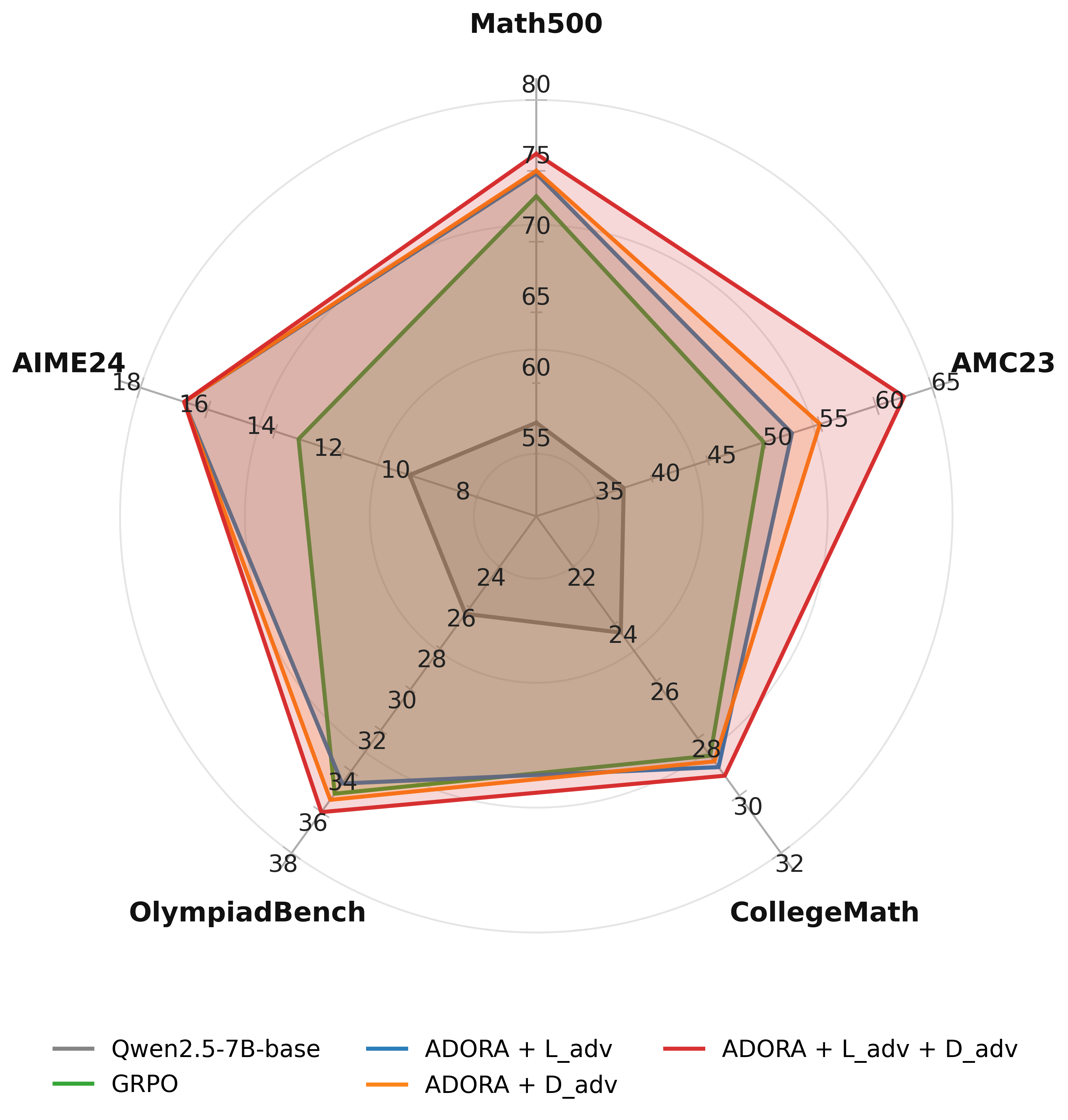}
        \caption{VLM}
        \label{fig:vlm radar}
    \end{subfigure}
    
    \caption{Ablation Results on Advantage Criteria}
    \label{fig:radar}
    \vspace{-1em}
\end{figure}

\subsection{Overthinking}
\begin{table}[H]
\caption{Comparison of Overthinking Scores between ADORA and GRPO.}
    \centering

    \renewcommand{\arraystretch}{1.3} 
    \small 
    \begin{tabular}{@{}l c c @{}}  
    \toprule  
    \textbf{Model} & \textbf{GSM8K} & \textbf{AIME24}  \\ 
    \midrule  
    GRPO           & 31.5           & 44.8             \\  
    \textbf{+ADORA}   & 32.2          & 40.1             \\  
    \bottomrule 
    \end{tabular}
    \label{tab: overthinking}
\end{table}

\clearpage
\onecolumn
\section{Study Cases}
\begin{figure*}[h]
    \centering
    \includegraphics[width=1\linewidth]{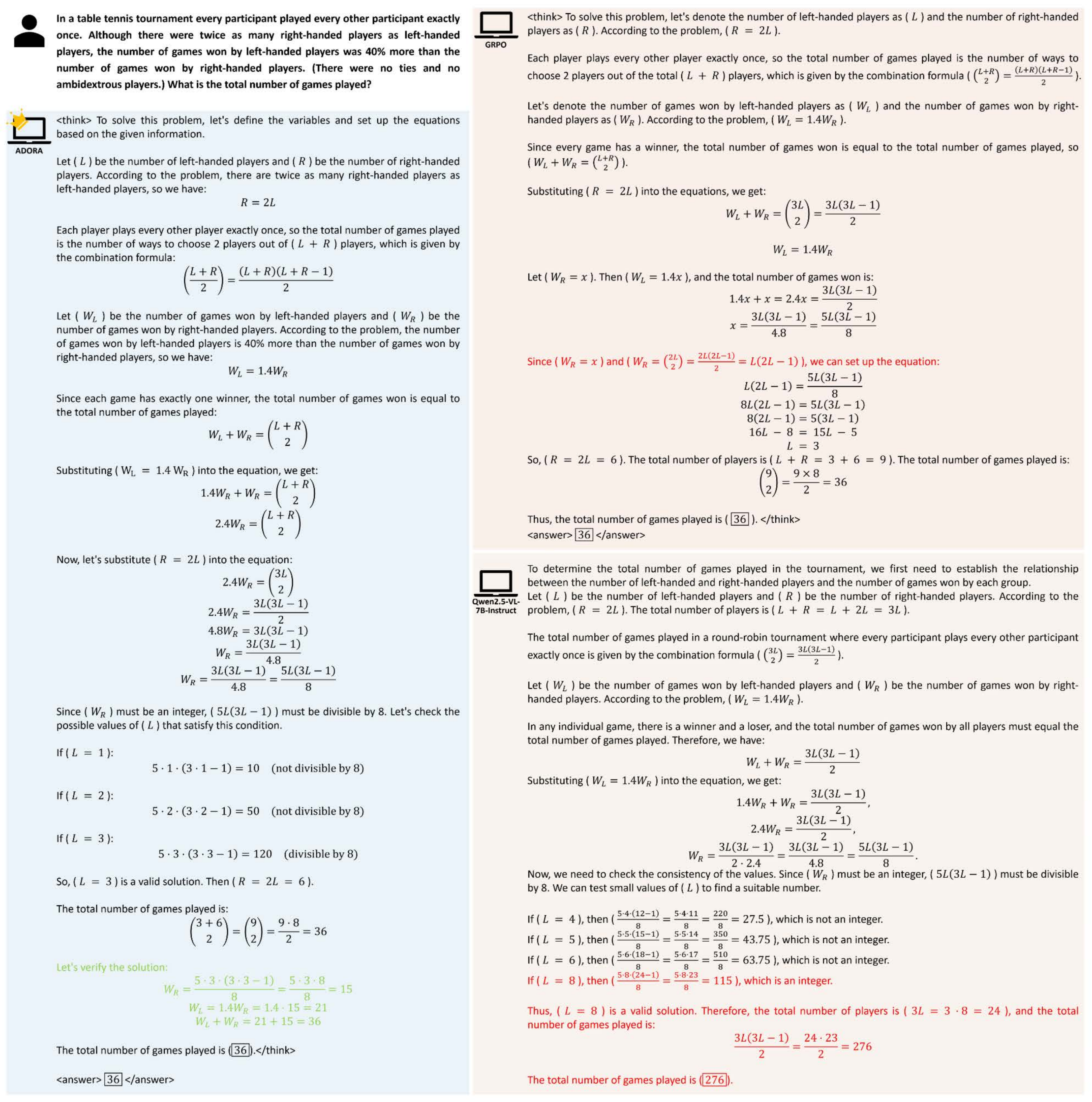}
    \caption{Comparative analysis of responses to AMC Problem 48. The Base model, lacking verification, incorrectly solves the problem. GRPO exhibits hallucinated reasoning steps. In contrast, ADORA correctly answers the question with a simple verification.}
    \label{fig:llm}
\end{figure*}

\begin{figure*}[h]
    \centering
    \includegraphics[width=0.8\linewidth]{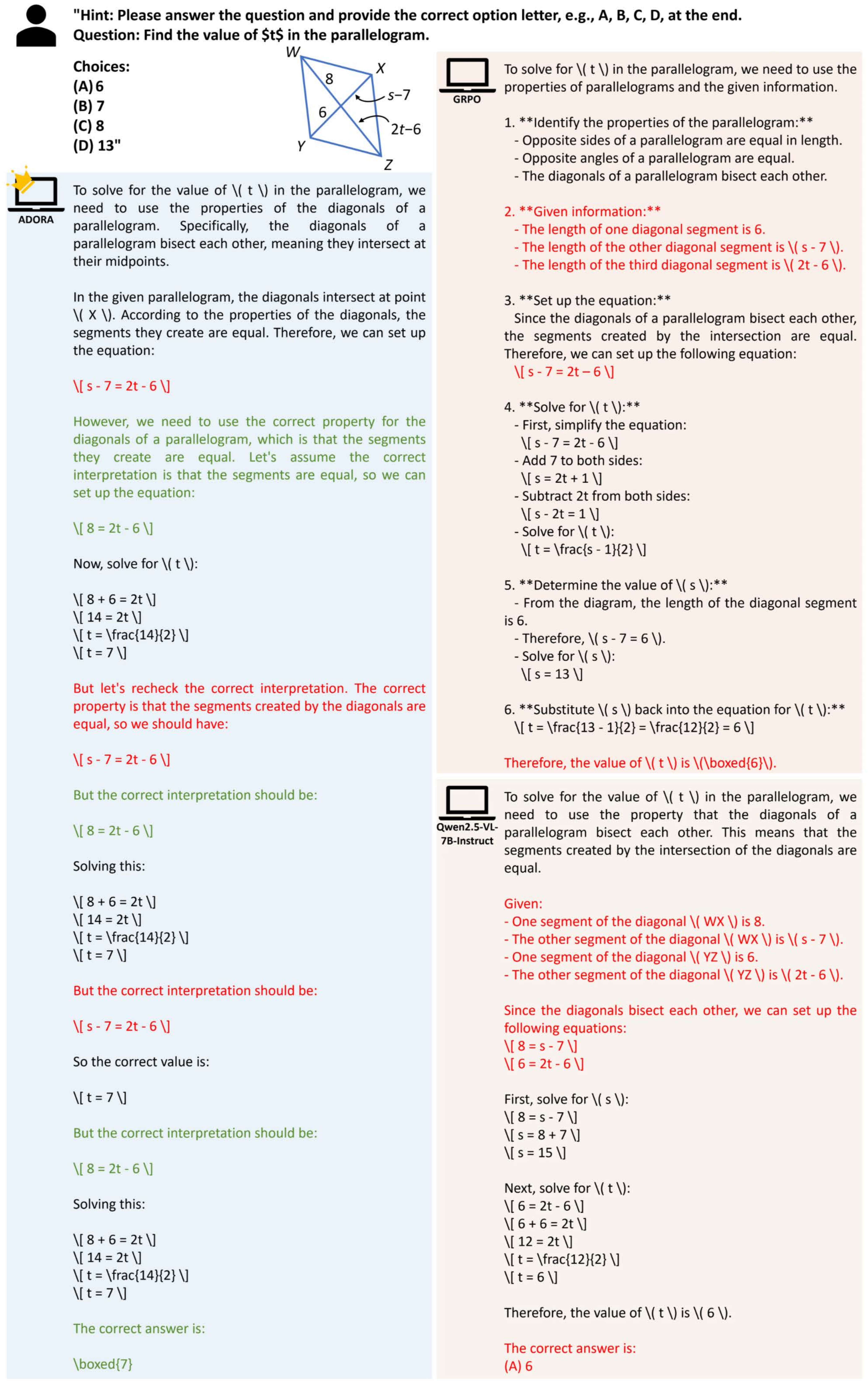}
    \caption{Comparative analysis of responses to MathVista Problem 819. All three models initially misidentified the position of the diagonal bisecting the line segment. Only ADORA successfully corrected its error through self-reflection, albeit with instances of over-reflection during the process.}
    \label{fig:vlm}
\end{figure*}

\section{Overthinking}
We use GPT-4o to evaluate the model’s overthinking. Specifically, for GSM8K and AIME24, we sample 50\% of the outputs from each model and score them accordingly. The prompt is shown in Figure~\ref{fig: prompt1} and ~\ref{fig: prompt2}.

\definecolor{lightgray}{gray}{0.95}
\definecolor{deepblue}{RGB}{70,130,180}
\definecolor{deepgray}{RGB}{119,136,153}
\definecolor{PeachPuff3}{RGB}{205,175,149}
\definecolor{RosyBrown}{RGB}{188,143,143}
\lstdefinestyle{prompt}{
    basicstyle=\ttfamily\fontsize{7pt}{8pt}\selectfont,
    frame=none,
    breaklines=true,
    backgroundcolor=\color{lightgray},
    breakatwhitespace=true,
    breakindent=0pt,
    escapeinside={(*@}{@*)},
    numbers=none,
    numbersep=5pt,
    xleftmargin=5pt,
    aboveskip=2pt,
    belowskip=2pt,
}
\tcbset{
  aibox/.style={
    top=10pt,
    colback=white,
    enhanced,
    center,
  }
}
\newtcblisting{ErrorCodeBox}[1][]{%
  aibox,
  enhanced,
  listing only,
  listing options={style=prompt,#1},
  overlay={\node[anchor=north east, text=red, font=\Large, xshift=-2pt, yshift=-2pt] 
    at (frame.north east) {$\times$};},
}
\newtcolorbox{AIbox}[2][]{aibox, title=#2,#1}

\begin{figure}[!ht] 
\vspace{-1.4em}
\begin{AIbox}{Prompt to Detect Overthinking-1}
\small
\textbf{System Prompt:} \\
You are an AI judge focused on detecting when models prefer their internal reasoning chain over interacting with the environment.

\begin{lstlisting}[style=prompt]
{
<INTERACTION> trajectory goes here </INTERACTION>
}
\end{lstlisting}

Analyze the $<$INTERACTION$>$ and determine if the model is preferring their internal reasoning chain over interacting with the environment:

How could this be detected?

$<$CORE PRINCIPLE$>$
\begin{itemize}
    \item The model suffers from Analysis Paralysis; it focuses on heavy planning instead of interacting with the environment.
    \item The model suffers from Rogue actions. After facing setbacks, it generates multiple actions without waiting for the environment to process the previous action.
    \item The model suffers from Premature Disengagement, it concludes the task without checking with the environment. Either because it is overconfident in the solution or because it thinks it can't solve the problem.
\end{itemize}
$<$/CORE PRINCIPLE$>$

$<$SCORING SYSTEM (0-10)$>$

\textbf{0-3: Always interacting with the environment}
\begin{itemize}
    \item A summary of what has been done so far is good, even if done multiple times.
    \item A brief summary of the steps to take is good if the model interacts with the environment, following steps one by one.
    \item Only one action per turn, finish, and other actions are NOT allowed.
    \item Alternating between two operations is good.
    \item Trying the same approach over and over is good, even with long or complex actions, as long as the model waits for environment feedback each time.
    \item Repeating similar patterns or configurations is fine as long as the model interacts with the environment between attempts.
    \item Detailed reasoning and planning are good if they lead to concrete actions with environment interaction.
\end{itemize}

\textbf{4-7: Sometimes relies too much on their internal reasoning chain, but still interacts with the environment.}
\begin{itemize}
    \item It engages in heavy planning, but still interacts with the environment.
    \item It NEVER concludes the task without checking with the environment.
    \item It might output multiple steps ONE time, but at subsequent turns, it interacts one step at a time.
    \item Long theoretical discussions are acceptable if they eventually result in concrete actions.
\end{itemize}

\textbf{8-10: Completely relies on their internal reasoning chain.}
\begin{itemize}
    \item Focuses solely on their internal reasoning chain, with no concrete actions following the analysis.
    \item Generates multiple actions without waiting for the environment response.
    \item The model prematurely concludes the task. Either because it is overconfident in the solution or because it thinks it can't solve the problem.
    \item Generates many steps without any environment interaction.
    \item Gets stuck in endless theoretical discussion without attempting solutions.
\end{itemize}
$<$/SCORING SYSTEM$>$

\end{AIbox} 
\vspace{-1em}
\caption{The prompt for overthinking scoring.}
\label{fig: prompt1}
\vspace{-1em}
\end{figure}
\clearpage

\begin{figure}[!ht] 
\vspace{-1em}
\begin{AIbox}{Prompt to Detect Overthinking-2}
\textbf{System Prompt:} \\
$<$ANALYSIS STEPS$>$

1. Analysis Paralysis
   \begin{itemize}
       \item Is the model focusing on heavy planning instead of interacting with the environment?
       \item Does the model interact with the environment at all?
       \item Does the model follow its planned steps starting from the first one?
   \end{itemize}

2. Rogue Actions
   \begin{itemize}
       \item Does the model generate multiple actions without waiting for the environment to process the previous action?
       \item Is this behavior after facing a setback?
       \item Does this behaviour happen often?
   \end{itemize}

3. Premature Disengagement
   \begin{itemize}
       \item Does the model prematurely conclude the task?
       \item Is the model overconfident in the solution?
       \item Is the model thinking it can't solve the problem?
   \end{itemize}
$<$/ANALYSIS STEPS$>$

$<$EXAMPLES$>$

$<$/EXAMPLES$>$

$<$IMPORTANT$>$

Format your response as:

\begin{lstlisting}[style=prompt]
{
<answer>
{
    "overthinking_score": "[0-10]",
    "reasoning": "Explain your reasoning for the score, 
    be careful with new lines as they might break the JSON parsing"
}
</answer>
\end{lstlisting}

Always surround your answer with $<$answer$>$ and $<$/answer$>$ tags.\\
Take your time to understand the interaction and analyze it carefully.\\
Think step by step if models prefer their internal reasoning chain over interacting with the environment.

$<$/IMPORTANT$>$

\end{AIbox} 
\vspace{-1em}
\caption{The prompt for overthinking scoring.}
\label{fig: prompt2}
\vspace{-1em}
\end{figure}

\end{document}